\documentclass[10pt,twocolumn,letterpaper]{article}

\usepackage[pagenumbers]{cvpr} 
\usepackage{tcolorbox}
\usepackage{threeparttable,graphicx,amssymb,textcomp,multirow,booktabs,siunitx,bold-extra,bold-extra,wrapfig,xspace}

\newcommand{\finding}[2]{
    \begin{tcolorbox}[
        colback=white!90!gray,     
        colframe=teal!60!black,     
        arc=5pt,                    
        boxsep=5pt,                 
        left=5pt,                  
        right=5pt,                 
        top=4pt,                    
        bottom=4pt,                 
        boxrule=0.8pt
    ]
    \vspace{-0.18cm}
        \paragraph{\textbf{\textit{Finding #1:}}} #2
    \vspace{-0.2cm}
    \end{tcolorbox}
}

\newcommand{\rot}[1]{\rotatebox{50}{#1}}

\usepackage{array}
\usepackage{graphicx}
\usepackage{amsmath}
\usepackage{amssymb}
\usepackage{booktabs}
\usepackage{multirow}
\usepackage{multicol}
\usepackage[table]{xcolor}%
\usepackage{bbding}

\newcommand{\myparagraph}[1]{\vspace{0.05cm}\noindent\textbf{#1}}
\newcommand{\mysubsubsection}[1]{%
  \par\addvspace{\medskipamount}%
  \noindent\textbf{#1}%
  \par\nobreak\noindent\ignorespaces
}

\definecolor{cvprblue}{rgb}{0.21,0.49,0.74}
\usepackage[pagebackref,breaklinks,colorlinks,allcolors=cvprblue]{hyperref}
\def\paperID{937} 
\def\confName{CVPR}
\def\confYear{2026}
\def\Ourbench{OpenBench\xspace}
\usepackage{pifont}
\definecolor{limegreen}{HTML}{32CD32}
\definecolor{deeper-gray}{gray}{0.70}
\definecolor{light-gray}{gray}{0.85}
\newcommand{\cmark}{\textcolor{limegreen}{\ding{51}}}
\newcommand{\xmark}{\textcolor{red}{\ding{55}}}
\def\eg{\emph{e.g.}}

\title{From Indoor to Open World: Revealing the Spatial Reasoning Gap in MLLMs}

\author{Mingrui Wu\textsuperscript{$1$}\qquad
Zhaozhi Wang\textsuperscript{$1$}\qquad
Fangjinhua Wang\textsuperscript{$2$}\\ \vspace{6pt}
Jiaolong Yang\textsuperscript{$3$}\qquad
Marc Pollefeys\textsuperscript{$2$}\qquad
Tong Zhang\textsuperscript{$1$}\textsuperscript{*}
\vspace{3pt}
\\
{\small
\textsuperscript{$1$}University of Chinese Academy of Sciences \qquad
\textsuperscript{$2$}ETH Z\"urich \qquad
\textsuperscript{$3$}Microsoft Research Asia
}
}

\begin{document}

\newcommand{\mycorrespondingauthor}{%
  \renewcommand{\thefootnote}{*}%
  \footnotetext{Corresponding author.}%
  \renewcommand{\thefootnote}{\arabic{footnote}}%
}

\twocolumn[{
    \renewcommand\twocolumn[1][]{#1}
    \maketitle
    
    \begin{center}
        \centering
        \captionsetup{type=figure}
        \includegraphics[width=\linewidth]{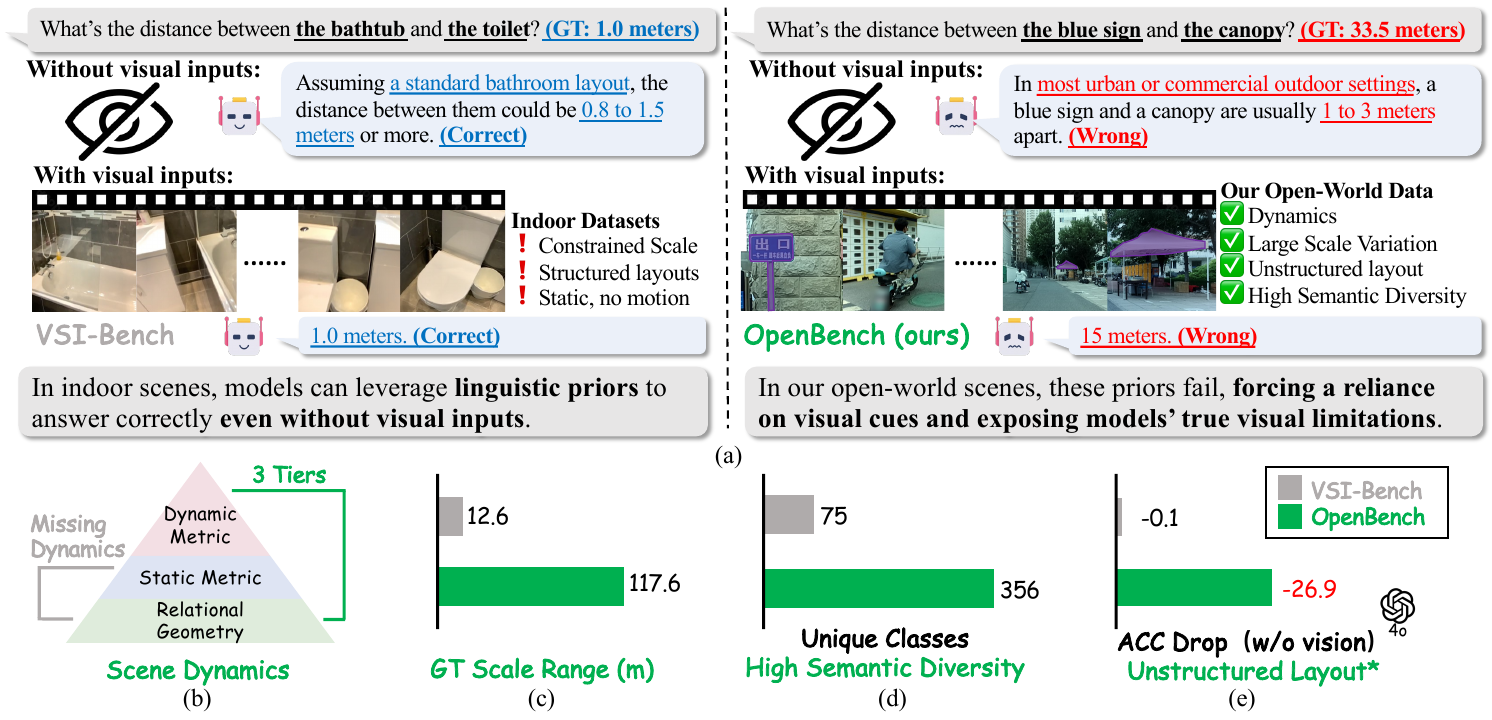}
        \vspace{-0.8cm}
        \caption{\textbf{Comparison between VSI-Bench~\cite{yang2025thinking} and OpenBench (ours).} \textbf{Top:} Qualitative comparison of Gemini-2.5-Pro's responses, with and without vision inputs, on VSI-Bench and OpenBench. \textbf{Bottom:} Quantitative comparison of two benchmarks in terms of evaluation coverage, scale range, semantic diversity, and accuracy drop without vision inputs. *The mean relative accuracy drop of GPT-4o on absolute distance tasks when vision inputs are disabled. A larger drop proves reliance on vision, not linguistic priors, in the unstructured layouts.} \label{fig:teaser}
        \vspace{-0.05cm}
    \end{center}
    }]

\mycorrespondingauthor

\begin{abstract}
While Multimodal Large Language Models (MLLMs) have achieved impressive performance on semantic tasks, their spatial intelligence—crucial for robust and grounded AI systems—remains underdeveloped. Existing benchmarks fall short of diagnosing this limitation: they either focus on overly simplified qualitative reasoning or rely on domain-specific indoor data, constrained by the lack of outdoor datasets with verifiable metric ground truth. To bridge this gap, we introduce a large-scale benchmark built from pedestrian-perspective videos captured with synchronized stereo cameras, LiDAR, and IMU/GPS sensors. This dataset provides metrically precise 3D information, enabling the automatic generation of spatial reasoning questions that span a hierarchical spectrum—from qualitative relational reasoning to quantitative metric and kinematic understanding. Evaluations reveal that the performance gains observed in structured indoor benchmarks vanish in open-world settings. Further analysis using synthetic abnormal scenes and blinding tests confirms that current MLLMs depend heavily on linguistic priors instead of grounded visual reasoning. Our benchmark thus provides a principled platform for diagnosing these limitations and advancing physically grounded spatial intelligence. Project page: \url{https://mingrui-wu.github.io/openbench/}
\end{abstract}

\vspace{-0.4cm}
\vspace{-0.3cm}
\section{Introduction}
\label{sec:intro}
\vspace{-0.1cm}

\begin{table*}[htbp]
    \centering
    \begin{threeparttable}
        \resizebox{\textwidth}{!}{
            \begin{tabular}{l c ccc l l}
                \toprule
                \textbf{Benchmark} & \textbf{Input Modality} & \textbf{Relational} & \textbf{Metric-Static} & \textbf{Metric-Dynamic}  & \textbf{Scene} & \textbf{Metric GT Source} \\
                \midrule
                OmniSpatial~\cite{jia2025omnispatial} & Single-view & \cmark & \xmark & \xmark & Internet Mixed & N/A \\
                SpatialBench~\cite{cai2024spatialbot} & Single-view & \cmark & \xmark & \xmark & Embodied & MDE$^{\dagger}$, only auxiliary \\
                SparBench~\cite{zhang2025sparbench} & Multi-view & \cmark & \xmark & \xmark & Indoor & N/A \\
                All-Angles-Bench~\cite{yeh2025seeingperspective} & Multi-view & \cmark & \xmark & \xmark & Indoor & N/A \\
                MMSI-Bench~\cite{yang2025mmsi} & Multi-view & \cmark & \xmark & \xmark & Internet Mixed & N/A \\
                SpaCE-10~\cite{gong2025space10} & Video + Pointcloud & \cmark & \xmark & \xmark & Indoor & Off-the-shelf Datasets~\cite{dehghan2021arkitscenes,dai2017scannet,yeshwanth2023scannetpp} \\
                \midrule
                SpatialRGPT-Bench~\cite{cheng2024spatialrgpt} & Single-view & \cmark & \cmark & \xmark & Internet & MDE$^{\dagger}$ \\ 
                Open3D-VQA~\cite{zhang2025open3dvqa} & Single-view & \cmark & \cmark & \xmark & UAV & MDE$^{\dagger}$ \\ 
                CA-VQA~\cite{daxberger2025mmspatial} & Multi-view & \cmark & \cmark & \xmark & Indoor & Off-the-shelf Datasets~\cite{dehghan2021arkitscenes} + MDE$^{\dagger}$ \\
                Ego3D-Bench~\cite{gholami2025ego3dbench} & Multi-view & \cmark & \cmark & \xmark & Autonomous & Off-the-shelf Datasets~\cite{nuscenes2019,sun2020waymo,chang2019argoverse} \\
                VSI-Bench~\cite{yang2025thinking} & Video & \cmark & \cmark & \xmark & Indoor & Off-the-shelf Datasets~\cite{dehghan2021arkitscenes,dai2017scannet,yeshwanth2023scannetpp} \\
                STI-Bench~\cite{li2025stibench} & Video & \cmark & \cmark & \cmark & Indoor\&Autonomous & Off-the-shelf Datasets~\cite{dai2017scannet,zhang2024omni6dpose,sun2020waymo} \\
                \midrule
                \textbf{\Ourbench(ours)} & \textbf{Video/Stereo Video} & \cmark & \cmark & \cmark & \textbf{Open-World} & \textbf{\parbox[c]{4.5cm}{ Metric accurate sensors \\ (Stereo, LiDAR \& IMU)}} \\
                \bottomrule
            \end{tabular}
        }
    \end{threeparttable}
    \caption{\textbf{Comparison of benchmarks for spatial intelligence evaluation.} Our benchmark is the first to provide metrically-sound ground truth for all three tiers of spatial intelligence in diverse scenarios. A parallel stereo-input version of the benchmark, featuring the same questions, is also provided to support future research. $^\dagger$ MDE: Monocular Depth Estimation.}
    \vspace{-0.4cm}
    \label{tab:benchmark_comparison}
\end{table*}

Multimodal Large Language Models (MLLMs)~\cite{alec2021clip,jeanbaptiste2022flamingo,li2024llavaov,zhang2024llavanextvideo,hurst2024gpt4o,team2023gemini,bai2025qwen25vl,Qwen3-VL,chen2024internvl,wang2025internvl35,lu2024ovis,Anthropic2024Claude,guo2025seed15vl} have emerged as pivotal tools for interpreting and reasoning about the visual world, achieving state-of-the-art performance in a wide range of semantic understanding tasks such as visual question answering~\cite{agrawal2016vqa} and image captioning~\cite{plummer2015flickr30k}. Building upon this success, recent research has sought to extend these models from semantic to spatial understanding—cultivating spatial intelligence~\cite{yang2025thinking}, a capability essential for physically grounded AI systems such as autonomous driving~\cite{wu2025generating} and embodied agents~\cite{Szot_2025_CVPR, li2024cogact}. 

Developing meaningful benchmarks is essential for accurately characterizing the spatial capabilities of MLLMs and guiding future advances. To provide a clear conceptual foundation, we formalize spatial intelligence as a three-level hierarchy of increasing complexity: 1) \textit{Relational Reasoning}, the qualitative understanding of spatial configurations including relative arrangement, orientation, and overall scene structure; 2) \textit{Metric Reasoning}, the quantitative estimation of absolute geometric properties such as distance, depth, and size; 3) \textit{Kinematic Reasoning}, the metric-aware understanding of scene dynamics, encompassing both object motion (\eg, velocity) and the observer’s ego-motion. 

This hierarchy helps to contextualize the state of existing evaluations. 
Nevertheless, most existing benchmarks~\cite{zhang2025sparbench, yeh2025seeingperspective, jia2025omnispatial} remain confined to qualitative relational reasoning, without addressing the core challenges of metric and kinematic understanding. Several recent efforts~\cite{yang2025thinking, li2025stibench} have advanced towards static quantitative evaluation, yet they are restricted to indoor scenes, where 3D meshes~\cite{dai2017scannet, yeshwanth2023scannetpp, dehghan2021arkitscenes} provide convenient ground truth. However, such scenes are limited in spatial scale, semantic diversity, and visual variability—rendering the evaluations unrepresentative of real-world conditions. Other attempts~\cite{cheng2024spatialrgpt} have explored outdoor scenarios using monocular images with pseudo-ground-truth depth derived from single-view estimation~\cite{yang2024depthanythingv2}, but these suffer from scale ambiguity~\cite{HartleyZisserman2004} and unreliable metric supervision. Compounding these issues, most MLLMs are pretrained on indoor or web-collected image datasets, raising the critical question: Do current models truly generalize their spatial knowledge to open-world environments?

To address these limitations, we construct a novel open-world benchmark designed to rigorously assess spatial intelligence under real-world conditions. Our dataset comprises high-resolution, pedestrian-perspective videos offering high diversity, covering over 200 distinct outdoor places. To ensure metrically precise supervision, data were collected using a multi-sensor rig combining synchronized \textit{stereo cameras}, \textit{high-precision LiDAR}, and \textit{IMU/GPS sensors}. 
Building on this, we introduce an automatic pipeline to extract spatio-temporal information and generate comprehensive question–answer evaluations spanning all three tiers of the spatial hierarchy: relational reasoning, metric reasoning, and crucially, kinematic reasoning.

\begin{figure*}[thb]
  \centering
  \includegraphics[width=\textwidth]{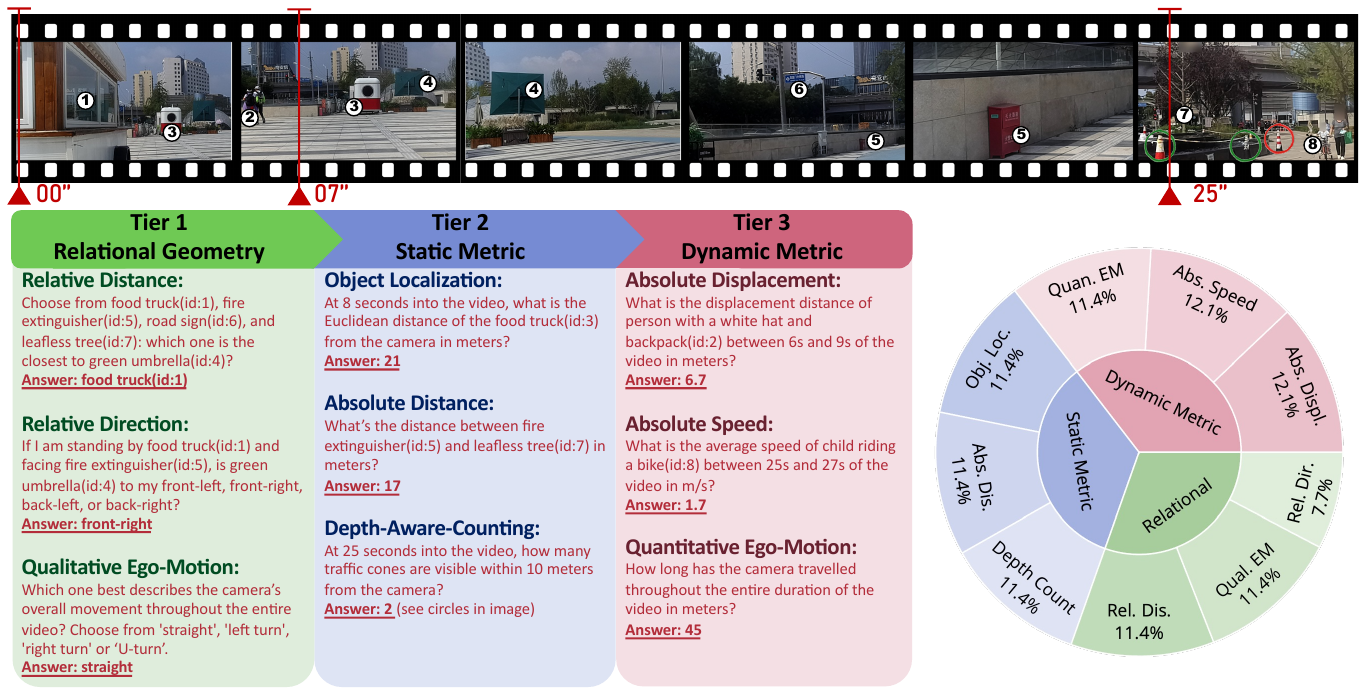}
  \caption{\textbf{Tasks and composition of our \Ourbench.} \textbf{(Left)} Representative question-answer examples for our three tiers. \textbf{(Right)} Illustration of the final task distribution, which is balanced across both tiers and tasks.}
  \label{fig:tasks}
  \vspace{-0.3cm}
\end{figure*}

Extensive evaluations of leading open- and closed-source MLLMs on \Ourbench yield conclusions that there are no evident improvement of spatial intelligence in open-world settings.
First, the metric reasoning ability of current MLLMs is largely superficial—driven by linguistic priors rather than genuine visual perception. As shown qualitatively in ~\cref{fig:teaser} (a) and quantitatively in ~\cref{fig:teaser} (e), our blinding test shows that the accuracy barely drops in the indoor VSI-Bench~\cite{yang2025thinking}, indicating limited usage of visual cues. Second, MLLMs fail to generalize the spatial knowledge acquired in constrained indoor environments to the complexities of the open world. Models that achieve large gains on indoor spatial benchmarks~\cite{yang2025thinking} show negligible improvement on \Ourbench, suggesting overfitting to indoor regularities rather than true spatial intelligence. Third, even the strongest models that approach human-level performance on static metric tasks collapse entirely on dynamic reasoning, revealing a fundamental deficiency in quantitative motion understanding. 

In a nutshell, our key contributions are summarized as:
\begin{itemize}
    \item \textbf{A benchmark and dataset for spatial intelligence}. We introduce \Ourbench, a metrically precise outdoor benchmark built from multi-sensor pedestrian-view data, with 8,736 question-answer pairs. We will also release additional raw multimodal videos.
    \item \textbf{Comprehensive evaluation of state-of-the-art (SoTA) MLLMs.} We conduct an extensive analysis of leading open- and closed-source models, offering the first unified assessment of spatial reasoning across static, relational, and dynamic tasks under real-world conditions.
    \item \textbf{Current spatial intelligence is fragile.} Our findings reveal that existing MLLMs lack generalizable spatial intelligence—their gains on indoor benchmarks do not transfer to open-world settings.
\end{itemize}


\section{Benchmark}
\vspace{-0.1cm}
\subsection{Overview}
\vspace{-0.1cm}
Unlike previous works~\cite{yang2025thinking,li2025stibench} that primarily repurpose off-the-shelf datasets, \Ourbench first establishes a purpose-built data foundation through a customized multi-sensor data collection effort that provides metrically-sound ground truth.
Second, we designed a pipeline that extracts spatial information and generates question-answer pairs covering the full spectrum of spatial intelligence, finalized by a careful human curation phase to ensure benchmark quality. 

\begin{figure*}[t]
  \centering
  \includegraphics[width=\textwidth]{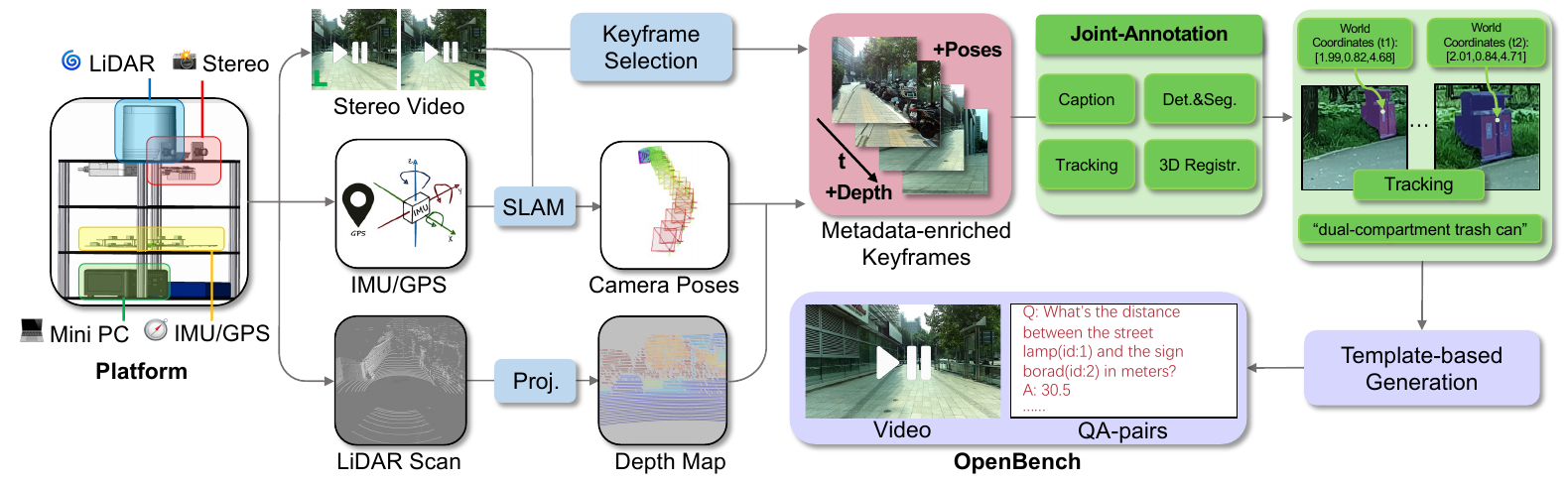}
  \vspace{-0.6cm}
  \caption{\textbf{Overview of our benchmark construction pipeline.}}
  \label{fig:pipeline}
  \vspace{-0.6cm}
\end{figure*}

The tasks in \Ourbench are structured as a hierarchy of three tiers, representing a progression of spatial reasoning capabilities. ~\cref{fig:tasks} provides a detailed breakdown of the task hierarchy and representative examples at each level. The hierarchy begins with \textbf{Relational Reasoning} (relative distance, relative direction, and qualitative ego-motion), which assesses a model's foundational ability to form a qualitative, non-metric spatial representation of the scene. 
The second tier, \textbf{Metric Reasoning}, comprising object localization, absolute distance, and depth-aware counting, introduces a more demanding challenge: grounding visual perception in an absolute metric scale to enable precise, quantitative estimation of static properties like distance. 
Finally, \textbf{Kinematic Reasoning} includes tasks such as absolute displacement and speed estimation. These require maintaining spatio-temporal consistency by tracking entities over time and understanding metric measurements and temporal intervals to compute dynamic quantities.

\subsection{Data collection}
\label{sec:data_collection}
\vspace{-0.1cm}

\myparagraph{Platform.} 
Our customized data collection platform, shown in the upper left of \cref{fig:pipeline}, is equipped with a stereo RGB camera system, a 32-beam LiDAR, and an IMU/GPS unit, all of which provide timestamp-synchronized data streams.
To ensure smooth motion, the platform is mounted on a manual cart navigated by walking operators.

\myparagraph{Data Processing and Curation.} 
Our collection resulted in terabytes of raw data, which were then processed to generate a multimodal dataset. First, stereo images, point clouds, and IMU/GPS data were extracted with timestamps and carefully calibrated. Then LiDAR points were projected to generate sparse depth maps. At last, all sequences were manually reviewed for quality, yielding approximately 20 hours of curated high-quality data.

\myparagraph{Scene Selection}. Our scene selection strategy is designed to ensure environmental diversity while maintaining a strictly pedestrian-centric perspective, not road-centric. To this end, our data collection spans a wide range of sites, including university campuses, public parks, open plazas, and historical sites. 
This pedestrian-centric perspective of open-world scenes, rarely represented in current MLLM evaluations, not only provides a critical complement to the field but also fulfills our objective of challenging linguistic priors in such unstructured environments.
Please refer to Sec. B in the supplementary material for details of data collection.

\subsection{Benchmark Construction}
\label{sec:benchmark_construction}
\vspace{-0.1cm}

We design a three-stage pipeline that transforms raw multimodal data into question–answer pairs. The stages include: (1) Data Preprocessing, (2) Spatial Information Extraction, and (3) QA Generation and Curation. An overview of the entire pipeline is illustrated in ~\cref{fig:pipeline}. Please refer to Sec. C in the supplementary material for more details and Sec. D for an error analysis of this pipeline.

\mysubsubsection{Stage 1: Data Preprocessing}
This initial stage transforms the raw multimodal recordings into a structured format for analysis. First, the data is segmented into shorter video clips, with all streams aligned by their timestamps. From this synchronized data, we generate two streams of geometric metadata: 1) a metric-scale camera pose for each frame is estimated using ORB-SLAM3~\cite{carlos2021orbslam3} from stereo images and IMU data, and 2) a sparse depth map is created by projecting the calibrated LiDAR point clouds onto the image plane. At last, we sample keyframes at a fixed interval from each clip and bundle them with their corresponding camera poses and depth maps. This process yields the \textbf{metadata-enriched keyframes} that serve as the primary input for the subsequent stage.

\mysubsubsection{Stage 2: Spatial Information Extraction}
We then employ a \textbf{Joint-Annotation Module} that integrates multiple expert models to extract semantic and 3D spatial information from the keyframes. A locally-run MLLM first identifies objects and generates textual descriptions, which guide detection and segmentation models to produce pixel-level masks. These masks are temporally tracked using a point-tracking model (e.g., CoTracker~\cite{karaev2024cotracker3}) for motion consistency. Each object's 3D point cloud is subsequently reconstructed from depth maps and registered in the world coordinate system using the estimated camera pose. Consequently, every object is represented by a structured spatio-temporal profile that combines a detailed caption with a continuous 3D trajectory.

\mysubsubsection{Stage 3: QA Generation and Curation}
Finally, we convert the extracted structured spatio-temporal information into the final, high-quality video-QA benchmark. Specifically, we first resolve object ambiguity in open-world scenes (e.g., multiple identical street lamps) by assigning each queried object a unique numerical ID, which is overlaid as a visual tag that consistently tracks the object throughout the video. All metric quantities (\eg, distance, speed) are then computed directly from the objects’ spatio-temporal coordinates. Using these annotations, we generate QA pairs through a template-based approach. The automatically generated QA pairs are subsequently refined via an MLLM-assisted, human-in-the-loop curation process. From this curated pool, we sample 1,000 video clips and nearly 1,000 verified QA pairs for each of the nine tasks, ensuring balanced coverage across the three reasoning tiers. The resulting benchmark, \Ourbench, comprises 8,736 high-quality QA pairs in total.

\section{Evaluation}
\vspace{-0.1cm}

\begin{figure*}[ht!]
\captionsetup{type=table}
\vspace{-0.4cm}
\centering
\fontsize{10.2pt}{10.0pt}\selectfont
\setlength\tabcolsep{4pt}
\renewcommand{\arraystretch}{1.0}
\scalebox{0.89}{
\begin{tabular}{r|cc|ccccccccc}
& & &
\rot{Rel. Dis.} &
\rot{Rel. Dir.} &
\rot{Qual. EM} &
\rot{Obj. Loc.} &
\rot{Abs. Dis.} &
\rot{Depth Count} &
\rot{Abs. Displ.} &
\rot{Abs. Speed} &
\rot{Quan. EM}
\\
Methods & Rank & Avg. & \multicolumn{3}{c}{\cellcolor{green!20}Relational $(\mathcal{MCA})$} & \multicolumn{3}{c}{\cellcolor{purple!20}Static Metric $(\mathcal{NA})$} & \multicolumn{3}{c}{\cellcolor{pink!20}Dynamic Metric $(\mathcal{NA})$}\\
\hline

\rowcolor{green!10}
\multicolumn{1}{l|}{\textcolor{black}{\textit{Against Human on tiny}}} & & & & & & & & & & & \\
Human-level & 1 & 60.3 & 85.7 & 83.3 & 73.7 & 43.9 & 39.2 & 67.5 & 42.9 & 65.8 & 66.8 \\
Gemini-2.5-Pro & 2 & 36.8 & 53.1 & 23.1 & 46.7 & 39.7 & 33.8 & 40.3 & 22.2 & 27.8 & 40.0 \\
GPT-5 & 4 & 27.9 & 37.5 & 30.8 & 40.0 & 35.3 & 25.3 & 12.8 & 9.2 & 31.4 & 33.0 \\
Qwen3VL-32B-Instruct & 3 & 31.9 & 56.3 & 23.1 & 33.3 & 26.2 & 10.9 & 32.8 & 14.4 & 37.2 & 52.3 \\
\hline

\rowcolor{green!10}
\multicolumn{1}{l|}{\textcolor{black}{\textit{Closed-source Models}}} & & & & & & & & & & & \\
Gemini-2.5-Pro & 1 & \cellcolor{deeper-gray}37.2 & \cellcolor{deeper-gray}50.0 & 28.1 & \cellcolor{deeper-gray}52.5 & \cellcolor{deeper-gray}37.4 & \cellcolor{deeper-gray}28.1 & 37.9 & \cellcolor{deeper-gray}26.8 & 31.1 & \cellcolor{light-gray}40.8 \\
Gemini-2.5-Flash & 6 & 19.5 & 17.9 & 2.8 & \cellcolor{light-gray}50.6 & 22.7 & 16.1 & 26.8 & 8.1 & 6.8 & 20.0 \\
GPT-5 & 2 & 29.7 & 34.4 & 33.1 & 49.5 & \cellcolor{light-gray}32.5 & 23.7 & 20.9 & 10.5 & 33.8 & 30.6\\
GPT-4o & 5 & 25.9 & 30.8 & 29.1 & 42.2 & 22.9 & \cellcolor{light-gray}27.0 & 21.6 & 17.5 & 15.5 & 28.8 \\
Claude-3.7-Sonnet & 4 & 26.5 & 38.9 & 32.8 & 47.6 & 31.3 & 22.4 & 31.5 & 5.2 & 30.1 & 5.0 \\
Doubao-Seed-1.6V & 3 & 27.3 & 35.9 & 24.1 & 44.0 & 16.6 & 18.9 & 38.7 & \cellcolor{light-gray}25.7 & 31.8 & 9.2 \\
\hline

\rowcolor{green!10}
\multicolumn{1}{l|}{\textcolor{black}{\textit{Open-source Models}}} & & & & & & & & & & & \\
InternVL2-8B & 14 & 24.5 & 35.1 & 31.7 & 40.8 & 21.8 & 17.8 & 39.7 & 15.0 & 17.8 & 3.8 \\
InternVL2-40B & 16 & 22.9 & 36.7 & 21.0 & 41.9 & 21.1 & 19.2 & 33.0 & 10.0 & 22.0 & 1.7 \\
InternVL3.5-2B & 18 & 21.7 & 34.8 & 32.1 & 40.1 & 3.8 & 2.7 & 40.8 & 11.5 & 16.6 & 17.0 \\
InternVL3.5-4B & 15 & 23.7 & 37.6 & 32.8 & 44.8 & 4.6 & 6.7 & 40.7 & 15.6 & 23.4 & 10.7 \\
InternVL3.5-8B & 4 & 28.5 & 37.6 & 33.6 & 47.3 & 12.2 & 13.2 & 42.3 & 20.3 & 30.2 & 21.5 \\
InternVL3.5-14B & 4 & 28.5 & 40.3 & 33.9 & 47.1 & 15.5 & 15.6 & \cellcolor{deeper-gray}42.8 & 21.8 & 32.1 & 9.0 \\
InternVL3.5-38B & 8 & 26.9 & 40.2 & 34.0 & 45.3 & 11.6 & 7.7 & \cellcolor{light-gray}42.7 & 20.3 & 31.4 & 11.1 \\
Qwen2.5VL-32B-Instruct & 3 & 30.0 & 41.7 & 32.7 & 44.4 & 23.9 & 24.0 & 27.7 & 16.7 & 32.8 & 27.3 \\
Qwen2.5VL-72B-Instruct & 11 & 26.5 & 38.5 & 20.1 & 46.5 & 27.7 & 26.0 & 29.4 & 9.4 & 29.7 & 9.8 \\
Qwen3VL-2B-Instruct & 21 & 18.4 & 33.7 & 32.4 & 37.5 & 2.2 & 4.2 & 22.4 & 6.6 & 19.0 & 12.5 \\
Qwen3VL-4B-Instruct & 19 & 21.1 & 34.8 & 23.7 & 46.4 & 3.9 & 6.9 & 24.1 & 13.6 & 20.4 & 17.5 \\
Qwen3VL-8B-Instruct & 2 & 31.2 & 38.3 & 31.2 & 49.3 & 21.0 & 15.1 & 33.3 & 21.3 & \cellcolor{light-gray}34.3 & 37.8 \\
Qwen3VL-32B-Instruct & 1 & \cellcolor{light-gray}32.2 & \cellcolor{light-gray}41.9 & 28.8 & 47.1 & 25.3 & 11.5 & 30.2 & 18.6 & \cellcolor{deeper-gray}36.8 & \cellcolor{deeper-gray}49.2 \\
LLaVA-OneVision-0.5B & 20 & 19.6 & 27.9 & 23.8 & 40.8 & 13.5 & 13.1 & 19.9 & 10.3 & 13.6 & 15.7 \\
LLaVA-OneVision-7B & 13 & 25.7 & 35.1 & 32.7 & 40.8 & 16.1 & 25.5 & 25.6 & 16.8 & 28.3 & 13.4 \\
LLaVA-OneVision-72B & 8 & 26.9 & 38.6 & 32.2 & 41.7 & 19.6 & 18.3 & 35.3 & 19.2 & 23.0 & 16.6 \\
LLaVA-Video-Qwen2-7B & 16 & 22.9 & 37.1 & 31.2 & 40.9 & 17.6 & 22.1 & 18.2 & 17.5 & 19.0 & 5.7 \\
LLaVA-Video-Qwen2-72B & 6 & 28.3 & 39.8 & 31.1 & 42.2 & 23.5 & 18.0 & 34.2 & 20.7 & 29.9 & 17.0 \\
Ovis2-4B & 12 & 25.9 & 34.6 & 30.3 & 40.1 & 16.1 & 18.6 & 36.8 & 17.7 & 22.6 & 18.7 \\
Ovis2-16B & 7 & 28.1 & 37.0 & \cellcolor{deeper-gray}35.8 & 42.0 & 20.0 & 6.2 & 41.7 & 21.7 & 23.3 & 28.2 \\
Ovis2-34B & 10 & 26.8 & 37.3 & \cellcolor{light-gray}35.5 & 40.8 & 19.1 & 13.1 & 37.5 & 18.7 & 27.4 & 15.5 \\
\hline
\end{tabular}
}
 \vspace{-0.3cm}
\caption{\textbf{Evaluation results for MLLMs.} We highlight the best and second-best results for each sub-task in \colorbox{deeper-gray}{deeper gray} and \colorbox{light-gray}{light gray}.}
\label{tab:main_table}
\vspace{-0.5cm}
\end{figure*}

\subsection{Evaluation Setup}
\vspace{-0.1cm}
\myparagraph{Benchmark Models.}
We evaluate a broad spectrum of MLLMs across various architectures and scales, covering both leading proprietary and publicly available models. All experiments are conducted using the VLMEvalKit~\cite{duan2024vlmevalkit} to ensure a standardized evaluation protocol.
Human performance is evaluated on a 270-question subset. For a fair comparison, we also report the results of several top-performing models on this same subset. 

\myparagraph{Evaluation Metrics.}
The tasks in \Ourbench are divided into two formats based on their expected answers: Multiple-Choice Answer (MCA) and Numerical Answer (NA). For MCA tasks, we report standard \textit{Accuracy}, and the random baseline is 25.0. For NA tasks, we primarily follow previous works~\cite{everingham2010pascal,yang2025thinking} and employ \textit{Mean Relative Accuracy} ($\mathcal{MRA}$). Please refer to Sec. E.1 and E.2 in the supplementary material for detailed models and metrics, and E.3 for human evaluation setup.

\subsection{Main Results}
\vspace{-0.1cm}
The main results of our evaluation are summarized in~\cref{tab:main_table}. All models perform far below human-level accuracy, highlighting a pervasive difficulty in reasoning about spatial relations and perceiving the world in a metrically grounded manner. Among all evaluated systems, the closed-source Gemini-2.5-Pro demonstrates a clear performance advantage, achieving the highest scores across five of the nine tasks. Within the open-source category, Qwen3-VL-32B stands out as the strongest model, achieving the best overall performance within this category.

Unlike prior indoor spatial benchmarks—such as All-Angles-Bench~\cite{yeh2025seeingperspective} and VSI-Bench~\cite{yang2025thinking}—which largely focus on recognition and linguistic reasoning, \Ourbench emphasizes the metric perception and reasoning by clearly dividing the overall tasks into 3 tiers. The following findings highlight key behaviors of current MLLMs under this setting, revealing their significant weaknesses in fine-grained metric reasoning.

\vspace{-0.1cm}
\finding{1}{Human-model disparity peaks on spatial relations rather than on metric estimation.}
Humans outperform models most dramatically on relational reasoning tasks (\eg, \textit{relative direction}: 83.3 vs. 23--30 for top models), while the performance gap narrows on metric estimation tasks (\eg, \textit{absolute distance}, \textit{depth-aware-counting}).
This indicates that what appears “intuitive” for humans—understanding relative spatial layouts—remains highly non-trivial for current MLLMs.
Unlike indoor benchmarks, which primarily expose perceptual or recognition deficiencies, \Ourbench surfaces a \textbf{cognitive gap} in relational reasoning, underscoring its diagnostic value for assessing true spatial understanding.
\vspace{-0.2cm}

\finding{2}{Performance on \Ourbench is not uniformly consistent; rather, models exhibit distinct performance profiles with clear strengths and weaknesses that are often specific to the model family.} 
Beyond the top-performing Gemini-2.5-Pro, we find that model performance lacks a clear correlation between overall scores and specific sub-task capabilities. For instance, while Qwen3VL-32B-Instruct is the leading open-source model on average, it performs poorly on specific tasks, ranking 18th out of 20 on \textit{relative direction} and 15th on \textit{absolute distance}. 
Moreover, we observe strong task specializations that appear to be inherent to model families. For instance, the OVIS family consistently excels at \textit{relative direction}, and the InternVL3.5 family demonstrates a significant and persistent advantage in \textit{depth-aware-counting}, with all its variants scoring above 40 on this task.

\vspace{-0.1cm}
\finding{3}{MLLMs fail to reason about quantitative spatial dynamics. While performance on static spatial tasks varies across models, Dynamic Metric reasoning remains a universal failure case.}
The top tier of spatial understanding involves estimating dynamic motion in the real world—a task requiring both metric precision and temporal consistency. Unlike static metric reasoning, where Gemini-2.5-Pro performs comparably to human annotators, all evaluated models exhibit a severe degradation when estimating motion-related quantities such as displacement or speed. 
This result reveals that current MLLMs lack a coherent representation of spatio-temporal continuity and struggle to integrate motion cues across frames. Rather than measuring an incremental weakness, this finding exposes a fundamental limitation of current vision-language architectures and highlights dynamic spatial reasoning as a critical next frontier for research.

\vspace{-0.1cm}
\finding{4}{The spatial intelligence observed on indoor benchmarks is largely a mirage—unable to transfer to open-world settings.}

\begin{figure*}[htbp] 
    \centering
    \includegraphics[width=\textwidth]{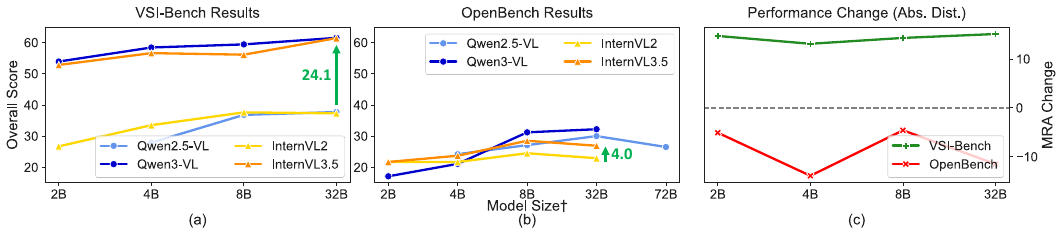} 
    \vspace{-0.85cm}
    \caption{\textbf{Performance comparison on VSI-Bench and \Ourbench across different sizes and model versions.} We evaluated models from two families, QwenVL and InternVL, each with an older and a newer version. All experiments used inputs of 32 frames for consistency.
    \textbf{(a)}\textbf{(b)}: the overall score of the models across various sizes on the indoor VSI-Bench and \Ourbench; lighter colors denote older model versions; the green arrow indicates the performance gain of InternVL3.5-38B over InternVL2-40B.
    \textbf{(c)}: the MRA change on the \textit{absolute distance} task when comparing InternVL3.5 to InternVL2. The green line highlights the performance gain on VSI-Bench, while the red line shows the performance drop on \Ourbench.
    \textdagger For plotting purposes, models are grouped by their approximate parameter scale. Refer to Sec. F.4 in the supplementary material for detailed results.
}
    \vspace{-0.4cm}
    \label{fig:scaling}
\end{figure*}

Multiple recent models~\cite{wang2025internvl35,Qwen3-VL} report dramatic performance gains on indoor spatial benchmarks like VSI-Bench~\cite{yang2025thinking} compared to their predecessors in the same model family. As shown in \cref{fig:scaling} (a), newer versions consistently outperform older ones by a substantial margin (e.g., +24.1 gain of InternVL3.5-38B to InternVL2-40B). As noted in InternVL3.5's technical report~\cite{wang2025internvl35}, more spatial QA data similar to the benchmarks are included, which we posit may explain the observed progress.
However, this progress fails to transfer to our open-world \Ourbench, as shown in \cref{fig:scaling} (b); newer model versions show only marginal performance gains over their predecessors.
Notably, on the \textit{absolute distance} task, which we adopt because its question templates are identical to those in VSI-Bench for a fair comparison and because measuring distance is a principled core ability of spatial perception, the newer InternVL3.5 models consistently underperform their InternVL2 predecessors (see \cref{fig:scaling} (c)). This provides more direct evidence that these generalist models are not acquiring generalizable spatial intelligence, but are instead overfitting to the statistical patterns of the indoor benchmarks.

\section{Linguistic priors or vision evidence?}
\label{sec: results_visionfails}
\vspace{-0.1cm}
After obtaining the main results, we perform additional ablation studies to uncover the underlying sources of failure. In this section, we first examine how current MLLMs rely heavily on linguistic priors rather than visual evidence when reasoning about spatial properties. Building on these observations, we then outline potential directions for developing models with more robust and explicit spatial perception.

\subsection{Linguistic priors}
\vspace{-0.1cm}
We posit that current general-purpose MLLMs fundamentally lack \textbf{metric spatial perception}. When presented with spatial tasks, they do not estimate geometric quantities—such as depth, distance, or physical size—directly from visual input. Instead, they approximate these values by drawing on linguistic priors and making semantic comparisons to familiar object categories. To validate the above hypothesis, we conduct two ablation studies below.

\myparagraph{Blinding Tests}. To examine the extent to which models genuinely rely on visual evidence, we evaluate them under a blinding setup where visual inputs are removed while the textual questions remain unchanged. As displayed in ~\cref{tab:blind}, most models exhibit only a marginal improvement (+2.2 to +6.3) when vision is enabled, even though our benchmark tasks are designed to heavily depend on visual information. By contrast, human performance increases substantially (+22.6) when vision is available, indicating a much stronger dependence on visual input. The limited gains on MLLMs suggest that MLLMs often rely on textual priors rather than truly grounded visual understanding. 
Overall, these results highlight a fundamental gap between human and model spatial reasoning.

\begin{table}[ht!]
\small
\sisetup{detect-all, explicit-sign}
\centering
\setlength\tabcolsep{1pt}
\renewcommand{\arraystretch}{1.0}
\scalebox{0.98}{
\begin{tabular}{l
    S[table-format = +2.1]  
    S[table-format = +2.1]
    S[table-format = +2.1]
    S[table-format = +2.1]
    S[table-format = +2.1]
    S[table-format = +2.1]
    S[table-format = +2.1]
    S[table-format = +2.1]
}
\toprule
Methods & {Avg} & {RDT} & {RDR} & {OL} & {ADT} & {DC} & {ADP} & {AS}\\
\midrule
\rowcolor{green!15}
Human$\dagger$ & +22.6 & +60.7 & +23.3 & +8.8 & +13.5 & +16.5 & +7.9 & +37.5 \\
Gemini & +12.4 & +15.3 & +0.9 & +17.7 & +16.2 & +24.3 & +2.6 & +7.2 \\
GPT & +2.2 & -3.3 & -1.2 & +9.3 & +8.5 & -0.2 & -13.6 & +15.1 \\
Qwen3-VL & +5.3 & +5.2 & +4.1 & +2.0 & +8.4 & +8.2 & -10.0 & +18.7 \\
InternVL3.5 & +6.3 & +0.6 & +1.0 & +11.2 & +7.7 & -0.3 & +14.5 & +7.3 \\
LLaVA-Video & +4.9 & +3.7 & +4.3 & +8.1 & +9.2 & -2.4 & +1.6 & +9.5 \\
LLaVA-OV & +3.3 & +4.0 & +11.0 & +8.0 & +12.5 & -4.0 & -1.9 & -3.5 \\
\bottomrule
\end{tabular}
}
\vspace{-0.1cm}
\caption{\textbf{Performance gain of vision-enabled over vision-disabled on \Ourbench}, evaluated on the largest or latest models within each model family. $^{\dagger}$ means evaluated on a tiny subset.}
\label{tab:blind}
\vspace{-0.2cm}
\end{table}

\begin{figure}[htbp] 
    \centering 
    \includegraphics[width=1\columnwidth]{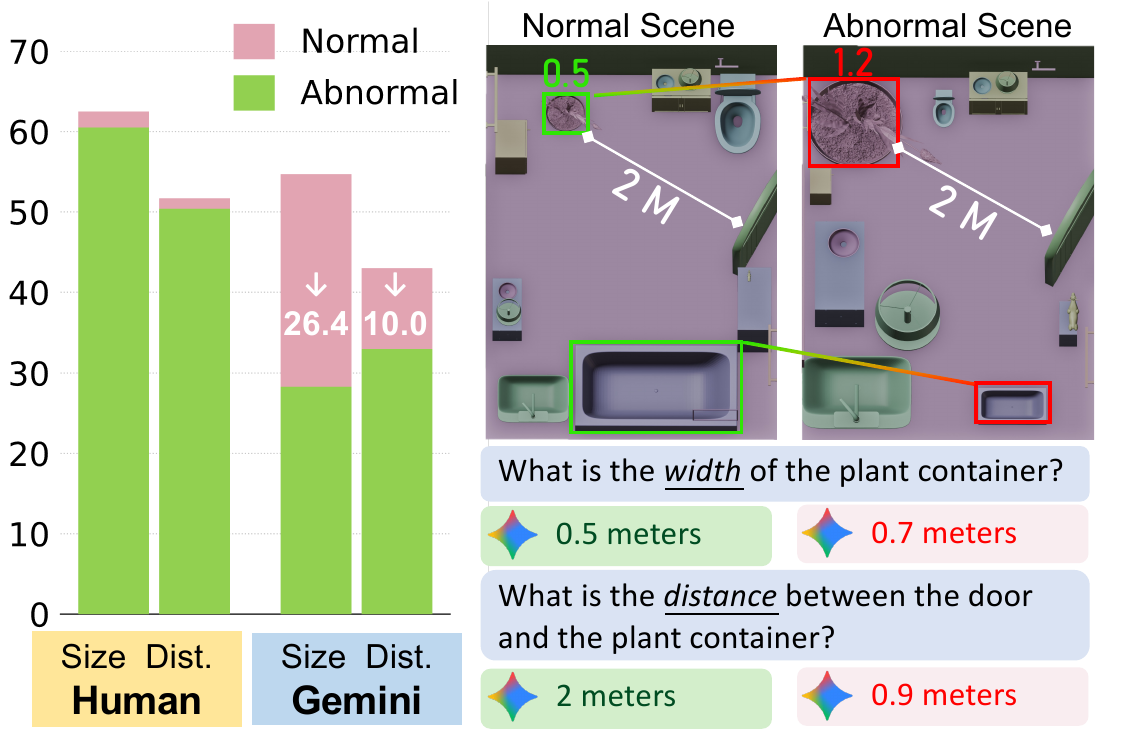} 
    \caption{\textbf{Illustrations and results of the synthetic test set.} The bar chart \textbf{(left)} shows the performance ($\mathcal{MRA}$) drop of humans and Gemini-2.5-Pro on the \textit{Size} and \textit{Distance} tasks when evaluated on abnormal scenes versus normal scenes.}
    \label{fig:synthetic}
    \vspace{-0.2cm}
\end{figure}

\myparagraph{Synthetic Abnormal Scenes.} To further investigate the extent of linguistic priors, we designed a controlled experiment using two sets of synthetic indoor scenes: a \textit{Normal Set} with conventional object proportions, and an \textit{Abnormal Set} in which object scales were deliberately manipulated while keeping the overall layout unchanged. We generated question–answer pairs for absolute distance and object size estimation on both sets. Results for human evaluators and Gemini-2.5-Pro are shown in~\cref{fig:synthetic}.

Gemini performance on both tasks degrades notably when evaluated on the \textit{Abnormal Set}. In contrast, human performance shows only a marginal degradation on these same tasks, indicating that the models’ spatial reasoning is strongly shaped by prior semantic knowledge: their metric estimates collapse when visual evidence contradicts familiar object statistics. More importantly, the drop is highly asymmetric. While accuracy decreases for the \textit{distance} task, the decline is far more severe for the \textit{size} task (e.g., from 54.7 to 28.3 for Gemini). This reflects the fact that size estimation is particularly vulnerable to class-level linguistic priors—models can bypass visual perception and simply output canonical object sizes, a strategy that fails catastrophically in abnormal scenes. The concurrent performance drop in the \textit{distance} task suggests that models also depend on these incorrect size estimates as reference cues for metric reasoning.

\subsection{Reasoning with geometric information}
\vspace{-0.1cm}
To identify the primary bottleneck in metric spatial tasks, we conduct an ablation study in which ground-truth geometric information is progressively revealed to the model. We focus on the \textit{absolute distance} task, which queries the distance between two objects that may not appear in the same frame. The distance $d$ can be calculated by:
\begin{equation} \label{eq:absdist}
    d = || (R \cdot p_2 + T) - p_1 ||,
\end{equation}
where $p_1$ and $p_2$ are the 3D positions of the two objects in their respective camera coordinate systems at times $t_1$ and $t_2$. The matrix $R$ and the vector $T$ represent the relative camera motion (rotation and translation) from $t_1$ to $t_2$. This formula decomposes the task into two main perceptual components: inferring object positions ($p_1, p_2$) and inferring camera ego-motion ($R, T$). 

\begin{table}[htbp]
    \centering
    \small
    \scalebox{0.9}{
    \begin{tabular}{
        l 
        S[table-format=2.1] 
        S[table-format=2.1]
    }
        \toprule
        \textbf{Setting} & {\textbf{Qwen3VL-32B}} & {\textbf{Gemini-2.5-Pro}} \\
        \midrule
        Vanilla                           & 17.5 & 19.2 \\
        + One Localization($p_1$)         & 19.2 & 27.9 \\
        + Both Localization($p_1$,$p_2$)  & 33.8 & 40.0 \\
        + Ego-Motion($R,T$)               & 32.5 & 22.9 \\
        + All($p_1, p_2, R, T$)           & 98.8 & 98.8 \\
        + All(w/o Formula)                & 59.2 & 85.4 \\
        \bottomrule
    \end{tabular}
    }
    \caption{Ablation study of progressively providing additional geometric information on the absolute distance task.}
    \label{tab:dist_ablation}
    \vspace{-0.2cm}
\end{table}

The results in~\cref{tab:dist_ablation} reveal several clear trends. First, the \textbf{Vanilla} setting performs even worse than in the main benchmark, suggesting that enforcing a structured multi-step calculation introduces additional parsing complexity that disrupts the model’s default reasoning strategy. Second, both models achieve near-perfect accuracy in the \textbf{All} setting, demonstrating that they can execute the mathematical computation without difficulty when all geometric quantities are provided. The sharp contrast between the \textbf{All} and \textbf{Vanilla} settings confirms that the dominant bottleneck is the extraction of precise metric information from the visual inputs, not the algebraic reasoning itself.

Furthermore, the substantial drop in the \textbf{All (w/o Formula)} condition indicates that models do not possess the 3D geometric knowledge required to derive the distance relationship independently—they act as reliable calculators only when explicitly instructed. Finally, although providing partial information (as in the \textbf{Both Localization} and \textbf{Ego-Motion} settings) yields moderate improvements, performance remains far from ideal, implying that models struggle with both object localization and ego-motion estimation when these components must be inferred visually.

\section{Related works}
\label{sec:related}
\vspace{-0.1cm}
\subsection{Multimodal Large Language Models}
\vspace{-0.1cm}
Multimodal Large Language Models (MLLMs) represent a significant advancement, extending the capabilities of Large Language Models (LLMs)~\cite{hurst2024gpt4o,touvron2023llama,deepseekr1} to jointly interpret and synthesize information across heterogeneous modalities~\cite{yin2024mllmsurvey}. The evolution of MLLMs from foundational methods like CLIP~\cite{alec2021clip} and BLIP-2~\cite{li2022blip,li2023blip2}, to pivotal architectures interfacing visual features with frozen LLMs like Flamingo~\cite{jeanbaptiste2022flamingo}, has led to contemporary systems (e.g., LLaVA~\cite{li2024llavaov,liu2024llavanext,zhang2024llavanextvideo}, Qwen-VL~\cite{bai2023qwenvl}, InternVL~\cite{chen2024internvl}) that achieve state-of-the-art performance on tasks like VQA~\cite{agrawal2016vqa} and scene understanding~\cite{azuma2022scanqa,plummer2015flickr30k}.

Despite this impressive progress, these foundational MLLMs continue to face significant challenges in visual-spatial reasoning~\cite{liu2023visual}. A strong reliance on textual priors often suppresses fine-grained spatial cues~\cite{ghatkesar2025looking}. This issue is exacerbated by the lack of compositional training data specifically designed to foster robust spatial intelligence. 

Separate from these foundational models, there is also a growing body of work~\cite{cai2024spatialbot,cheng2024spatialrgpt,cheng2025sr3d,gholami2025ego3dbench,wang2025videoanchor} focused on enhancing specific spatial reasoning capabilities. While these specialized models excel within their designated domains, they typically rely on task-specific training, auxiliary geometric inputs (e.g., depth maps, 3D reconstructions), or explicit region-level supervision. As a result, their capabilities are closely tied to the training distribution and do not reflect the innate spatial understanding of a general-purpose model. Moreover, many such methods are built on static scenes, limiting their ability to reason about motion or dynamic spatial relationships. In contrast, our benchmark evaluates the foundational spatial intelligence of general-purpose MLLMs without introducing new supervision or additional geometric modalities.

\vspace{-0.1cm}
\subsection{Benchmarking Spatial Intelligence}
\vspace{-0.1cm}
Early efforts to evaluate the spatial reasoning abilities of MLLMs primarily focused on reasoning within static single-image scenes, thus assessing qualitative relational understanding~\cite{johnson2017clevr,tang2024sparkle, shiri2024empirical,stogiannidis2025mind,santhosh2025doesspatial, yao2025lens,wang2025spatial457,liu2025can,dongfang2025multimodal}.
Recently, approaches have evolved to incorporate multi-view images~\cite{yeh2025seeingperspective,yang2025mmsi,gholami2025ego3dbench} to capture scene complexity and viewpoint changes, or video inputs~\cite{zhou2025vlm4d,yang2025thinking,zhang2025do,gong2025space10} to evaluate temporal-spatial capabilities. 

Beyond qualitative assessment, works like SpatialRGPT~\cite{cheng2024spatialrgpt} and VSI-Bench~\cite{yang2025thinking} began evaluating quantitative measurements in static scenes, based on image and video inputs, respectively. Aiming to reduce linguistic bias, SIRI-Bench~\cite{song2025siri} uses synthetic videos generated by rendering 3D geometry math problems. Further extending this, STI-Bench~\cite{li2025stibench} introduced the quantitative evaluation of dynamic quantities like speed and displacement. 

Nevertheless, the domain imbalance across existing benchmarks remains a major limitation. Most datasets focus on constrained indoor scenes with limited viewpoint diversity and lack coverage of open-world and pedestrian-centric settings. Such bias makes it difficult to assess genuine improvements in spatial reasoning. Moreover, current benchmarks often test recognition and retrieval rather than higher-order reasoning about spatial composition, causality, or physical consistency—key aspects of spatial AI.

\vspace{-0.2cm}
\section{Conclusion and discussion}
\label{sec: conclusion}
\vspace{-0.1cm}
\myparagraph{Conclusion}. We introduced \Ourbench, an open-world benchmark for evaluating spatial intelligence in MLLMs across relational, metric, and kinematic reasoning. Built from pedestrian-perspective videos with precise geometric supervision, it enables realistic assessment beyond indoor or synthetic datasets. Our findings expose a structural gap between today’s MLLMs and the level of spatial understanding required for physically grounded AI. They further suggest that scaling visual encoders or expanding training corpora alone is insufficient; genuine progress will require mechanisms capable of inferring, storing, and manipulating 3D geometric quantities in a principled manner.

\myparagraph{Discussion.} Our ablation studies point toward several promising research avenues. Improving metric depth perception in open-world settings is essential, as this emerges as the principal bottleneck across spatial tasks. Integrating explicit geometric representations—whether through 3D-aware architectures, multi-view consistency, or hybrid symbolic–neural reasoning—may help models move beyond surface-level priors. Moreover, dynamic reasoning remains notably underexplored. Developing models that maintain temporally consistent world models, reason over motion trajectories, and utilize sensor-derived ego-motion data represents a critical next step.

\clearpage
{
    \small
    \bibliographystyle{ieeenat_fullname}
    \bibliography{main}
}

\clearpage
\setcounter{page}{1}
\setcounter{section}{0}
\renewcommand\thesection{\Alph{section}}
\renewcommand\thesubsection{\thesection.\arabic{subsection}}

\maketitlesupplementary

\section*{Appendix Outline}

In the supplementary material, we provide:
\begin{itemize}
    \setlength\itemsep{0em}
    \item Visual illustration of linguistic priors;
    \item Technical details of data collection and benchmark construction;
    \item Error analysis and pipeline validation;
    \item Evaluation setups and experiment details;
    \item More evaluation results.
    \item Privacy statement.
\end{itemize}

\section{Illustration of Linguistic Priors}
\begin{figure}[htbp] 
    \centering 
    \includegraphics[width=1\columnwidth]{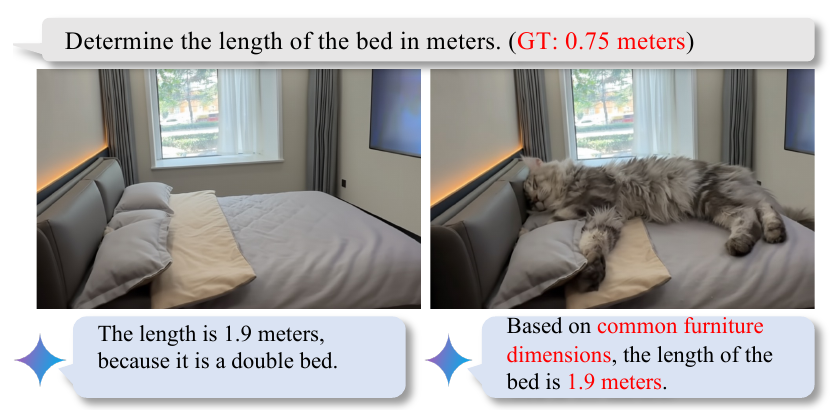} 
    \vspace{-0.8cm}
    \caption{\textbf{Illustration of a Prior-Driven Reasoning Failure.} The scene is a 1:3 scale miniature bed. \textbf{(Left)} Gemini-2.5-Pro defaults to its internal knowledge, identifying the object as a ``double bed" and outputting a prior-based estimate of 1.9m. \textbf{(Right)}Even when a strong, contradictory visual clue (a real cat) is introduced, the model fails to ground its reasoning in the visual evidence, and still defaults to the same ``common furniture dimension" prior and outputs 1.9m.}
    \label{fig:linguistic_prior}
\end{figure}

To clearly demonstrate that models rely on \textbf{linguistic priors} when answering spatial questions, and that this reliance can lead to wrong answers, we use a scene from a miniature room to ask models about size questions, as illustrated in \cref{fig:linguistic_prior}.

\section{Details of Data Collection}
\subsection{Hardware Specifications}
Our custom-built data collection platform, illustrated in the main paper, is equipped with a multi-sensor suite designed for high-fidelity, pedestrian-centric data capture. The core components include:
\begin{itemize}
    \item A synchronized stereo RGB camera system (rolling shutter, 1080p, 15 FPS).
    \item A 32-beam omnidirectional LiDAR (10 FPS).
    \item A high-frequency Inertial Measurement Unit (IMU) operating at 100 Hz.
    \item A GPS unit operating at 1 Hz.
\end{itemize}
All sensors are connected to an onboard Intel NUC mini PC running Ubuntu, which is orchestrated by the widely-adopted ROS2 framework. This system provides unified, single-command control over the entire sensor suite and logs all multimodal data streams as timestamp-synchronized ROS2 bag files, facilitating robust subsequent extraction and processing. The entire platform is powered by an onboard battery, which provides approximately one hour of continuous operation per charge.

To ensure smooth motion capture, the platform was mounted on a manual cart. The camera system was positioned at approximately 1.4 meters from the ground. This height was deliberately chosen as a trade-off to balance the need for a pedestrian-like perspective with the critical requirement of maintaining motion stability during collection.

\subsection{Calibration and Rectification}
A rigorous multi-sensor calibration pipeline was executed prior to all data processing to ensure the metric accuracy and spatio-temporal alignment of our dataset. This process was divided into three main components:

\myparagraph{Stereo Camera Calibration and Rectification.}
We first calibrated the stereo camera system. This process involved using a standard chessboard pattern with the OpenCV library~\cite{opencv_library} to precisely determine the intrinsic parameters of each camera and the extrinsic transformation between the left and right camera units. Following calibration, a stereo rectification algorithm, also from OpenCV, was applied to all image pairs. This step is critical as it warps the two images such that their epipolar lines become collinear and horizontal, which allows for subsequent stereo matching and SLAM tasks. All visual data used in the downstream modules of our pipeline consists of these rectified stereo images.

\myparagraph{LiDAR-to-Camera Calibration.}
To fuse visual and depth information, we calibrated the extrinsic parameters (the 6-DoF transformation) between the LiDAR and the left camera of the stereo system. This calibration was performed using the OpenCalib ToolBox~\cite{opencalib}, which provides robust automatic calibration. The resulting transformation matrix is essential for our pipeline, as it allows us to accurately project the 3D LiDAR point clouds onto the 2D image plane of the camera to generate the metric-scale sparse depth maps.

\myparagraph{IMU Calibration.}
Finally, the Inertial Measurement Unit (IMU) was calibrated in two stages. We first determined its intrinsic parameters using the Kalibr toolbox~\cite{rehder2016kalibr}. Subsequently, the extrinsic transformation between the IMU and the stereo camera system was computed using the OpenCalib ToolBox~\cite{opencalib}. This precise IMU-camera extrinsic calibration is a prerequisite for ensuring the accuracy and robustness of the stereo-inertial SLAM module used for camera pose estimation.

\subsection{Operators and Scene Selection}
Operators were instructed to navigate the cart through the selected scenes at a typical walking speed. Turns were intentionally included in the routes to better represent natural pedestrian movement and to generate diverse trajectories for the final evaluation tasks. The entire data collection process totaled over 100 person-hours.

\begin{figure*}
    \centering
    \includegraphics[width=1\linewidth]{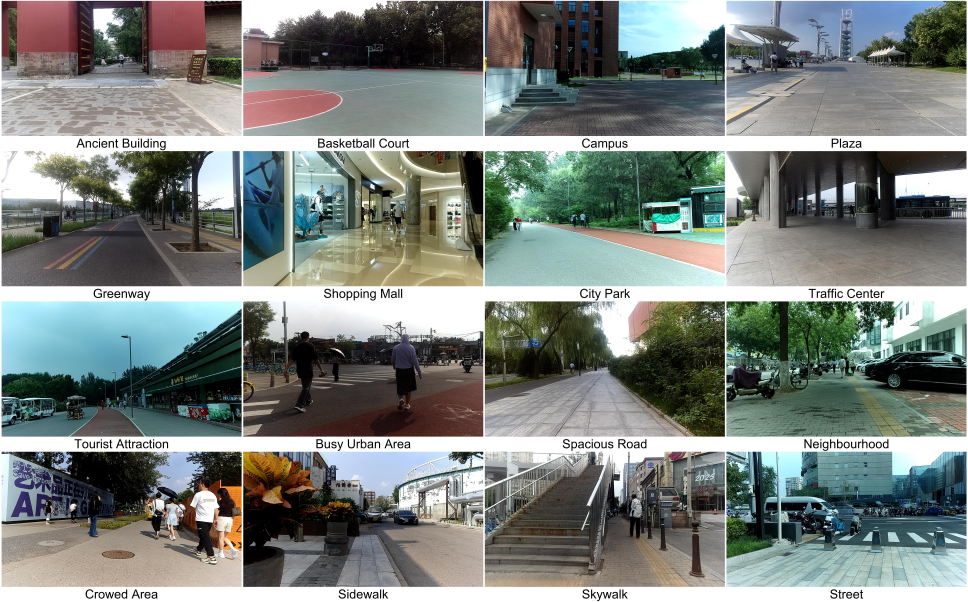}
    \caption{\textbf{Representative Samples of Scenes in \Ourbench.}}
    \label{fig:data_samples}
\end{figure*}

As referenced in the main paper, our scene selection strategy focused on maximizing diversity while capturing typical environments a pedestrian encounters. Furthermore, other than outdoor scenes, we include large-scale shopping malls to incorporate complex indoor scenarios. Unlike the confined residential or office environments common in existing benchmarks, large malls feature open layouts and a significantly larger range of scales, presenting open-world spatial challenges. Representative samples of our chosen scenes are shown in ~\cref{fig:data_samples}. 
We recognize that open-world environments can be more spacious and semantically sparse compared to object-dense indoor scenes. Therefore, our collection protocol intentionally prioritized pedestrian-centric areas known for a high density of potential query targets (e.g., street furniture, signage, complex storefronts, and other pedestrians) to ensure the final benchmark is both diverse and challenging.

Following a manual curation process, where we discarded sequences with poor visual quality (e.g., blur, low light) or those lacking distinct queryable objects, we obtained a 20-hour high-quality multimodal dataset. While approximately 6 hours of this data were used for the benchmark construction detailed in this paper, the remaining dataset, including all raw sensor data, will be made publicly available to the research community to foster further development in spatial intelligence.

\section{Details for \Ourbench construction}
\label{sec:benchmark_detail}

\myparagraph{Video Segmentation and Synchronization.}
For processing efficiency, we segment the long-form raw recordings \textbf{logically} rather than physically. Instead of duplicating or moving source data, each clip is defined by a \texttt{JSON} file that bundles all necessary, timestamp-synchronized metadata, including paths to the stereo image frames, LiDAR PCD files, and IMU/GPS data. These logical clips are defined with randomized durations between 15 to 30 seconds. This duration, shorter than those commonly used in indoor datasets~\cite{dai2017scannet}, is a deliberate design choice for two reasons: (1) the high semantic density of open-world scenes ensures sufficient complexity, comparable to much longer indoor recordings; and (2) it aligns with the sparse-frame sampling approach inherent to video loading for most MLLMs, ensuring effective capture of dynamic events.

\myparagraph{Camera Pose Estimation.}
We employ ORB-SLAM3~\cite{carlos2021orbslam3} in its stereo-inertial mode, which utilizes our rectified 15 FPS stereo image sequences and IMU data. We found that processing the original full-resolution (1080p) frames was unreliable for this algorithm. Therefore, all input images were first downsampled to 960x540, and their corresponding camera intrinsic parameters were rescaled accordingly. This process yields a precise, metric-scale pose in a world coordinate system for every frame.
This method was selected after an empirical evaluation on our data, where it demonstrated superior robustness and accuracy compared to the classical SfM pipeline COLMAP~\cite{schoenberger2016sfm,schoenberger2016mvs}.

\myparagraph{Densified Depth Map Generation.}
To address the inherent sparsity of single-frame LiDAR scans, we leverage the estimated camera poses to perform multi-frame point cloud fusion. 
For each frame, point clouds from temporally adjacent frames are transformed and aggregated according to their relative poses, yielding a denser fused point cloud. This fused representation is then projected onto the image plane—following the projection protocol described earlier—to produce a densified depth map. These enhanced depth maps are essential for accurate downstream 3D information extraction. In our final implementation, we set the temporal fusion window to 3, meaning the point cloud for each frame is aggregated with those of its immediately preceding and succeeding frames.

\myparagraph{Keyframe Selection.}
We extract keyframes from each video clip at a fixed 30-frame interval. This sampling rate (equivalent to 2 seconds at 15 FPS) was chosen to strike a balance: it is frequent enough to capture the vast majority of objects appearing in the pedestrian-speed video, yet sparse enough to avoid redundant annotation efforts on highly similar, consecutive frames. 

\myparagraph{Captioning with MLLMs.}
For captioning objects in the keyframes, our initial approach adopted the pipeline from GroundedSAM~\cite{ren2024groundedsam}, using an open-vocabulary tagging model~\cite{zhang2023recognize} to generate candidate class names. However, we observed that for our pedestrian-centric, open-world scenes, these models exhibit a strong bias towards labeling large, semantically dominant regions. This resulted in a high density of unsuitable, non-object labels such as `road', `sky', or `buildings', which are ill-suited as query targets for a spatial benchmark. Thus, we choose to instruct a locally-run MLLM, Qwen-2.5-VL-8B-Instruct~\cite{bai2025qwen25vl} to identify multiple semantically clear, physically distinct objects in each keyframe. For each object, the model generates a detailed caption describing only its intrinsic properties (\eg, `a red fire hydrant') and classifies it as either static or dynamic. This yields high-quality, relevant textual descriptions for subsequent grounding. See ~\cref{fig:prompts_caption} for the prompt used in this phase.

\begin{figure*}
\begin{tcolorbox}[colback=black!5!white,colframe=black!75!black,title=Captioner Prompt]

[Task]\\
Your task is to analyze an image and output a JSON object containing lists of captions forstatic and dynamic objects.

[Rules]\\
1. Content Principle: Identify distinct physical objects. Each string in the lists must describe a **single entity**. Avoid plural or group descriptions (e.g., instead of ``cars", list ``blue sedan", ``white SUV").\\
2. Ignore: Backgrounds (walls, roads), 2D elements (text, signs), minor parts of larger objects, natural or amorphous categories like ``tree", ``bush", ``flowers", ``grass", or ``cloud".\\
3. Uniqueness: If an object has already been listed, do not list it again. Once no new unique objects can be found, STOP immediately and close the JSON list with a bracket.\\
4. Classification: The identified objects should first be classified into static or dynamic.\\
   static objects: Stationary items. (e.g., a parked car, a bench, a trash bin).\\
   dynamic objects: Items visibly in motion. (e.g., a person walking, a car driving, a bird flying).\\
5. Captions: All captions must be purely visual descriptions (e.g., ``red sports car", ``person in yellow jacket"). Do not describe motion (e.g., avoid ``person walking").\\
6. Empty Lists: If no objects of a certain type are found, use an empty list ``[]".\\

[Example]
\begin{verbatim}
{
  "static_objects": [
    "red sports car",
    "black street lamp",
    "green park bench"
  ],
  "dynamic_objects": [
    "man in blue jacket"
  ]
}
\end{verbatim}

Your entire response must be a single, valid JSON object, strictly following the format shown in the `[example]`. Do not add any other text.

\end{tcolorbox}
\caption{\textbf{Prompt for the MLLM captioner.}}
\label{fig:prompts_caption}
\end{figure*}

\myparagraph{Object Detection and Segmentation.}
The object captions generated in the previous step serve as text prompts for GroundingDINO~\cite{liu2023groundingdino} to produce 2D bounding boxes. These boxes are then refined into pixel-level segmentation masks by the SAM model~\cite{kirillov2023sam}. 
As each caption corresponds to a single object instance, we select only the highest-confidence detection per caption, followed by Non-Maximum Suppression (NMS) to resolve potential part-whole ambiguities.

\myparagraph{Temporal Object Tracking.}
To ensure robust temporal consistency, we employ a point-tracking paradigm over a mask-propagation approach. While recent segmentation models like SAM2~\cite{ravi2024sam2segmentimages} offer tracking, their propagation-based method can lead to drift and errors in long-term tracking. 
We use a point tracking model (CoTracker3~\cite{karaev2024cotracker3}) to establish motion correspondence. We sample points within an object's keyframe mask, track them bidirectionally, and then use these tracked points on non-key frames with SAM to generate a final, temporally coherent mask sequence.

\myparagraph{3D Spatial Registration.}
To obtain each object's 3D representation, we use the per-frame mask to extract the corresponding depth values from the depth map. These values are then used to deproject the masked pixels into a 3D point cloud, whose centroid serves as the object's estimated 3D position relative to the camera. By combining this relative position with the frame's global camera pose, we register all objects into a unified world coordinate system. We note that as our LiDAR data primarily captures the front-facing surfaces of objects, we do not provide estimations of their full 3D dimensions (width, height, depth) in the current version of the benchmark.
A final de-duplication step is performed in this world space to merge instances of the same static object detected in different keyframes. 

\begin{figure*}[p]
    \centering
    \vfill
    \includegraphics[width=0.97\linewidth]{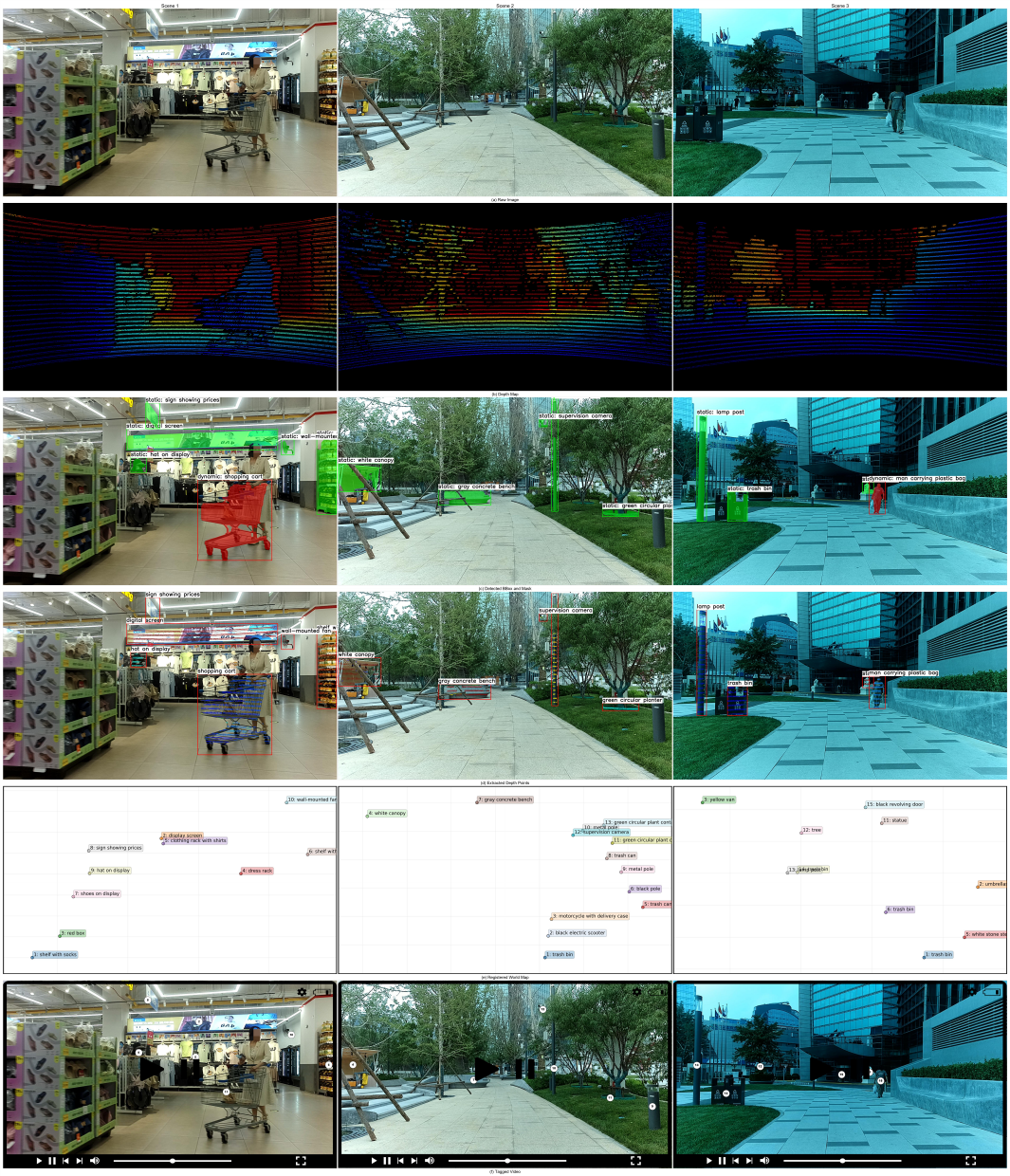}
    \vspace{-10pt}
    \caption{\textbf{Samples showing the workflow of the Joint-Annotation Module.} In the actual implementation, multiple keyframes are processed for each scene.}
    \label{fig:annotation}
\end{figure*}

See ~\cref{fig:annotation} for detailed samples illustrating the Joint-Annotation Module's workflow.

\myparagraph{Template-based Generation.}
We use a template-based approach to generate all question-answer pairs in order to separate and test the models' spatial reasoning capabilities, minimizing the influence of complex language understanding or multi-step logical reasoning as confounding variables. The full templates for 9 tasks are shown in ~\cref{tab:templates}.

\begin{table*}[h]
\centering
\begin{tabular}{p{4cm} p{10cm}}
\hline
\textbf{Task} & \textbf{Question Template} \\
\hline
Relative Distance & Measuring from the closest point of each object, which of the following is closest to the \textcolor{red}{\{Q\_class\}}(id:\textcolor{red}{\{Q\_id\}}): \textcolor{red}{\{A\_class\}}(id:\textcolor{red}{\{A\_id\}}), \textcolor{red}{\{B\_class\}}(id:\textcolor{red}{\{B\_id\}}), \textcolor{red}{\{C\_class\}}(id:\textcolor{red}{\{C\_id\}}), or \textcolor{red}{\{D\_class\}}(id:\textcolor{red}{\{D\_id\}})? \\
\hline
Relative Direction & If I am standing by the \textcolor{red}{\{C\_class\}}(id:\textcolor{red}{\{C\_id\}}) and facing the \textcolor{red}{\{A\_class\}}(id:\textcolor{red}{\{A\_id\}}), is the \textcolor{red}{\{B\_class\}}(id:\textcolor{red}{\{B\_id\}}) to my front-left, front-right, back-left, or back-right? The directions refer to the quadrants of a Cartesian plane. \\
\hline
Qualitative Ego-Motion & Assuming the video is recorded from a first-person perspective, which of the provided options best describes the person's overall movement throughout the entire duration of the video? Choose from straight, left turn, right turn or U turn. \\
\hline
Object 3D Localization & At approximately \textcolor{red}{\{T\}} seconds into the video, what is the Euclidean distance of the \textcolor{red}{\{A\_class\}}(id:\textcolor{red}{\{A\_id\}}) from the camera in meters? \\
\hline
Absolute Distance & What's the distance between the center of the \textcolor{red}{\{A\_class\}}(id:\textcolor{red}{\{A\_id\}}) and the \textcolor{red}{\{B\_class\}}(id:\textcolor{red}{\{B\_id\}}) in meters? \\
\hline
Depth-aware Counting & At approximately \textcolor{red}{\{time\_sec\}}s, how many \textcolor{red}{\{class\_name\}}s are visible within \textcolor{red}{\{distance\_threshold\}} meters from the camera? \\
\hline
Absolute Displacement & What is the displacement distance of the \textcolor{red}{\{A\_class\}}(id: \textcolor{red}{\{A\_id\}}) between \textcolor{red}{\{T1\}}s and \textcolor{red}{\{T2\}}s in meters? \\
\hline
Absolute Speed & What is the average speed of the \textcolor{red}{\{A\_class\}}(id: \textcolor{red}{\{A\_id\}}) between \textcolor{red}{\{T1\}}s and \textcolor{red}{\{T2\}}s in m/s? \\
\hline
Quantitative Ego-Motion & How long has the camera travelled throughout the entire duration of the video in meters? \\
\hline
\end{tabular}
\caption{\textbf{Templates used for question-answer pairs generation.}}
\label{tab:templates}
\end{table*}

\myparagraph{MLLM-assisted Curation.}
First, a powerful closed-source MLLM (e.g., Gemini-2.5-Pro~\cite{team2023gemini}) performs an initial pass to correct inaccurate captions and filter out questions related to objects that are poorly visible due to occlusion or detection errors. Second, the MLLM assigns a confidence score to the remaining QA pairs, flagging those with potential ambiguities or unstable visual tags. Finally, human annotators conduct a final review of all low-confidence samples, either correcting or discarding them.

\begin{figure*}
    \centering
    \includegraphics[width=1\linewidth]{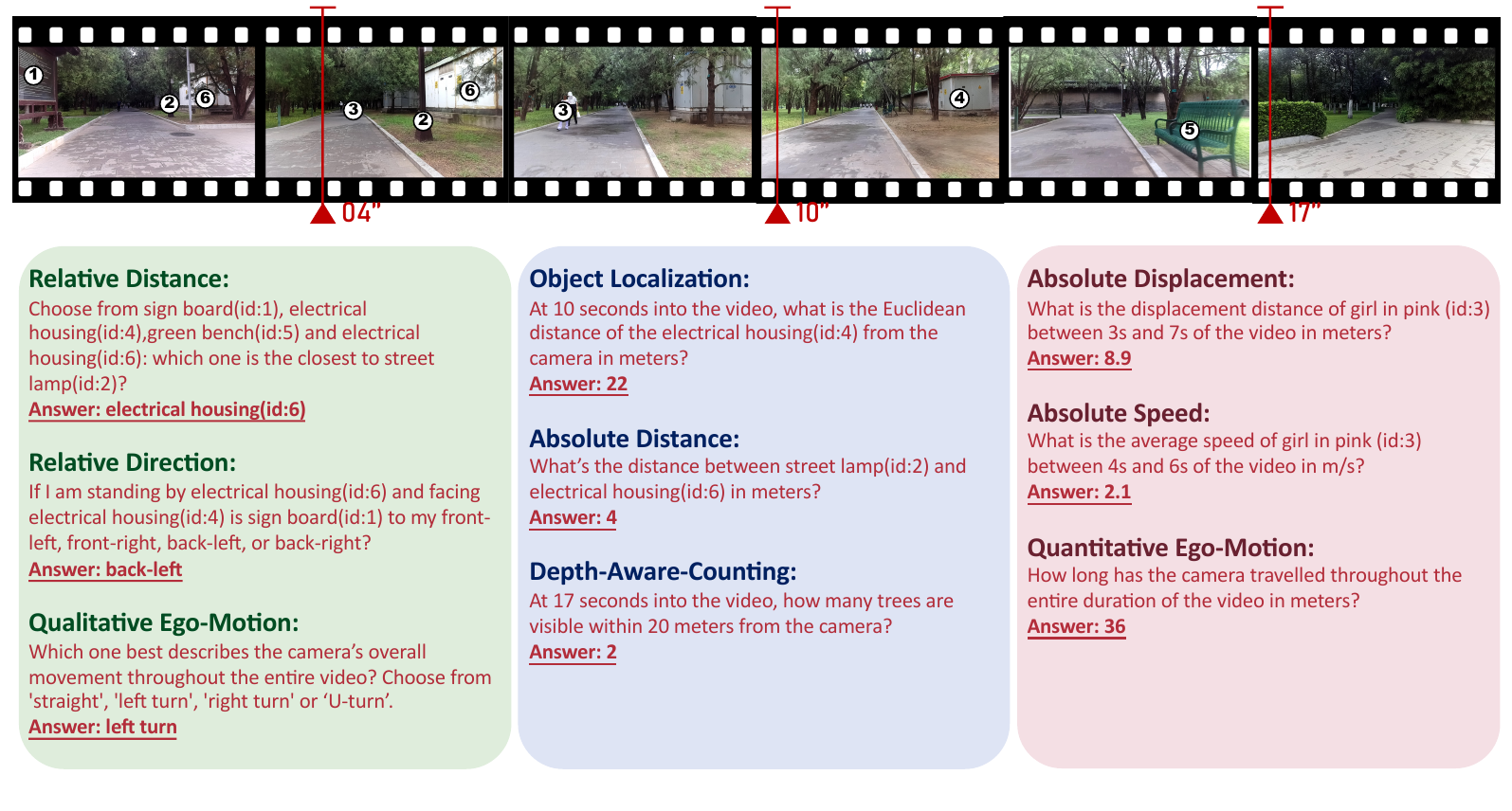}
    \caption{\textbf{\Ourbench examples.}(Part 1)}
    \label{fig:task-1}
\end{figure*}

\begin{figure*}
    \centering
    \includegraphics[width=1\linewidth]{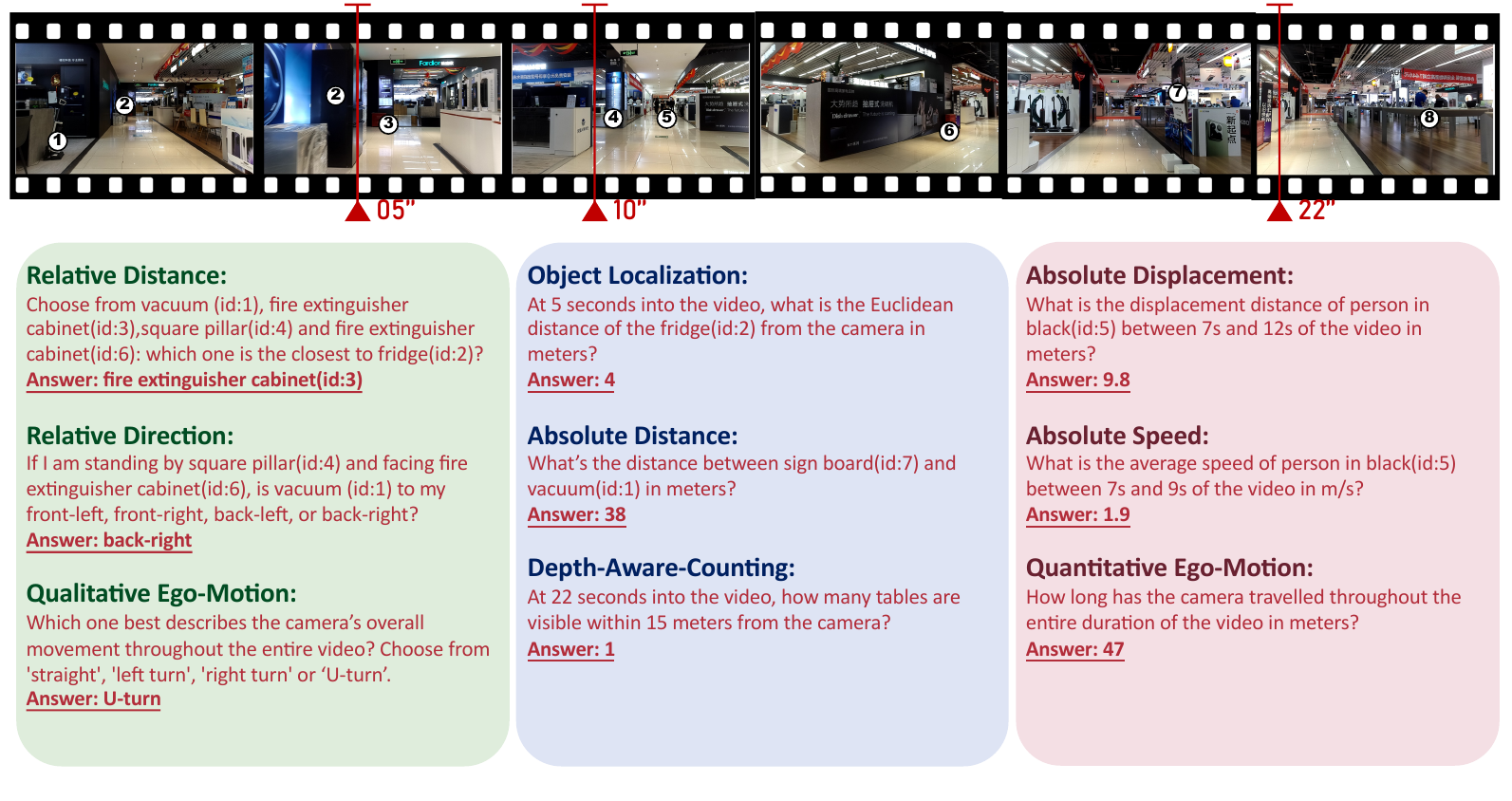}
    \caption{\textbf{\Ourbench examples.}(Part 2)}
    \label{fig:task-2}
\end{figure*}

\begin{figure*}
    \centering
    \includegraphics[width=1\linewidth]{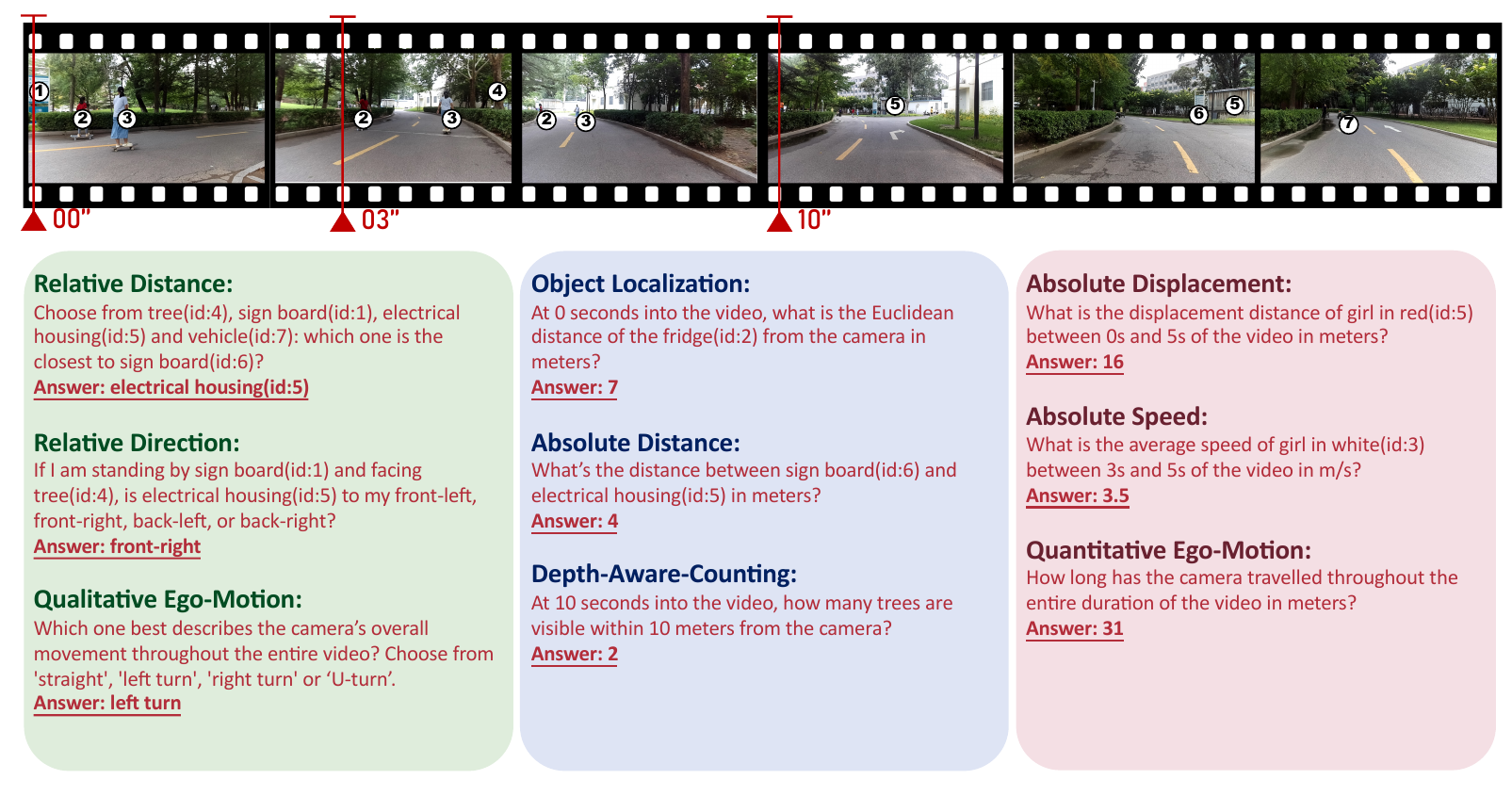}
    \caption{\textbf{\Ourbench examples.}(Part 3)}
    \label{fig:task-3}
\end{figure*}

See ~\cref{fig:task-1}, \cref{fig:task-2} and \cref{fig:task-3} for more examples in \Ourbench.

\section{Error Analysis of Benchmark Construction}
\label{sec:appendix_error_analysis}
Unlike benchmarks built upon existing, manually annotated 3D datasets~\cite{yang2025thinking,li2025stibench,gholami2025ego3dbench}, \Ourbench employs a highly automated pipeline to extract spatial information and generate QA pairs. While this automation enables scalability, it is crucial to analyze the potential sources of error. In this section, we analyze these errors and demonstrate that they are minimal, and that the quality of \Ourbench is ensured through our rigorous calibration, validation, and curation processes.

Three primary aspects could introduce errors into the final ground truth answers: (i) the extrinsic and intrinsic calibration of the sensors, (ii) the pose estimation from the SLAM algorithm, and (iii) the final 3D registration of objects in the world coordinate.

\subsection{Errors in Calibrations}
To ensure data fidelity, we conducted several independent calibration processes for all sensors and selected the optimal results.
The quality of the stereo calibration can be measured by its \textit{reprojection error}, which quantifies the distance between a detected pattern keypoint and its corresponding point projected from the other camera view. Using a standard chessboard pattern, our final selected calibration (from 25 image pairs) achieved a mean reprojection error of \textbf{0.32 pixels}.

Similarly, the LiDAR-to-camera calibration quality was measured by the reprojection error between 3D LiDAR points and their corresponding 2D image keypoints. Using 25 pairs, the mean reprojection error was \textbf{0.51 pixels}. We further validated this by measuring the planarity error between the checkerboard plane in the LiDAR point cloud and in the camera view. The mean translation error was \textbf{0.002 meters} and the mean rotation error was \textbf{0.978 degrees}, indicating a highly accurate spatial alignment between the sensors.

\subsection{Errors in SLAM Pose Estimation}
A quantitative evaluation of the final pose accuracy from ORB-SLAM3~\cite{carlos2021orbslam3} on our dataset is not feasible due to the lack of ground-truth trajectories in our collection. Instead, we rely on the extensive public validation of the algorithm itself. The original ORB-SLAM3 paper, for example, reports an Absolute Trajectory Error (ATE) of \textbf{0.035 meters} on comparable stereo-inertial datasets, demonstrating its high metric precision.

\subsection{Errors in World 3D Registration}
Several potential error sources could introduce error to the final 3D registration of semantic objects. These include visual occlusions, imperfect segmentation masks, minor temporal misalignments between the keyframe and the fused depth map, and the approximation of an object's center using its visible point cloud centroid. 

\begin{figure}[t]
    \centering
    \begin{subfigure}{\linewidth}
        \centering
        \includegraphics[width=\linewidth]{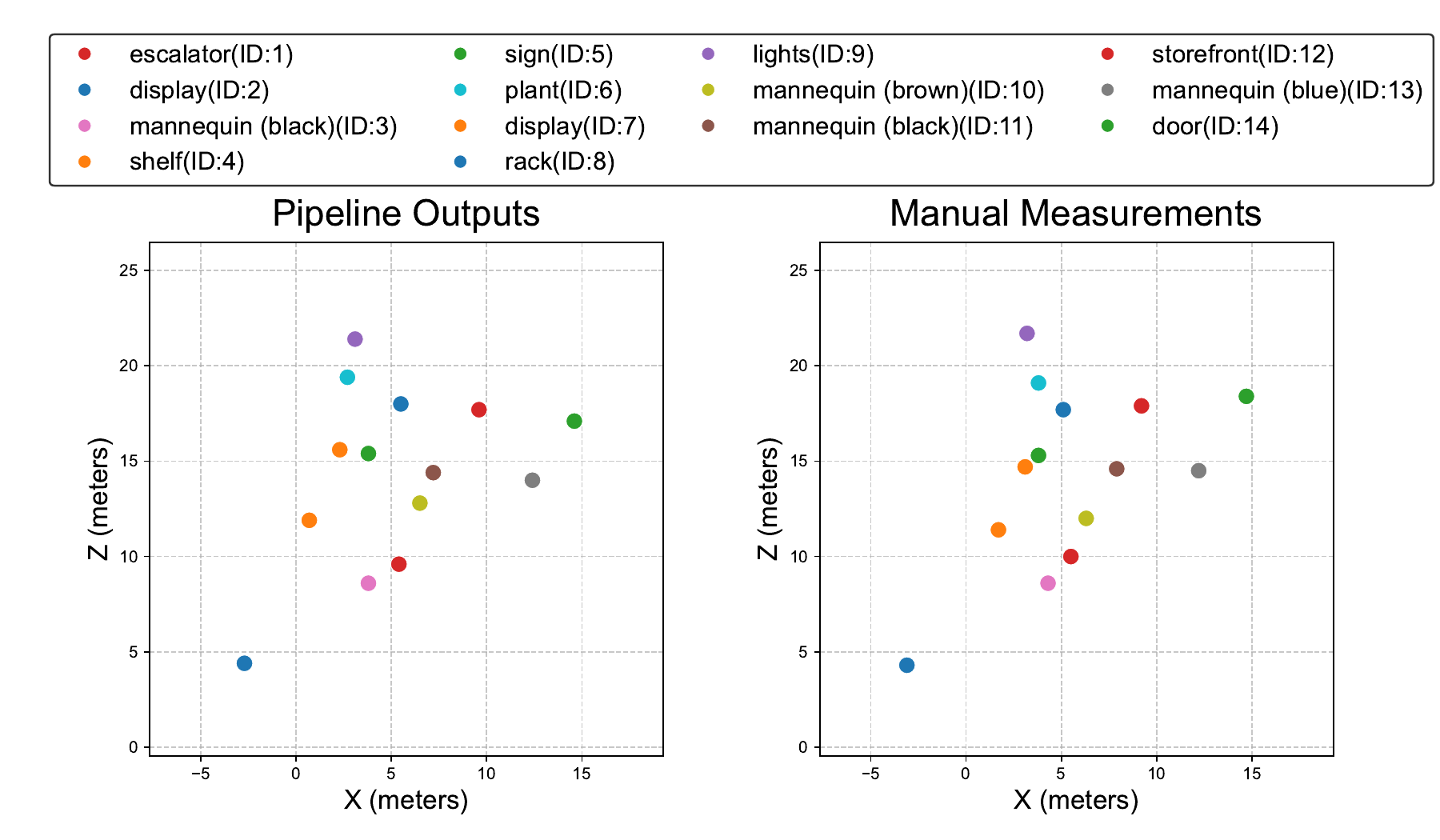}
        \caption{Validation of our pipeline in an indoor mall scene.}
    \end{subfigure}
    \vspace{4pt}
    \begin{subfigure}{\linewidth}
        \centering
        \includegraphics[width=\linewidth]{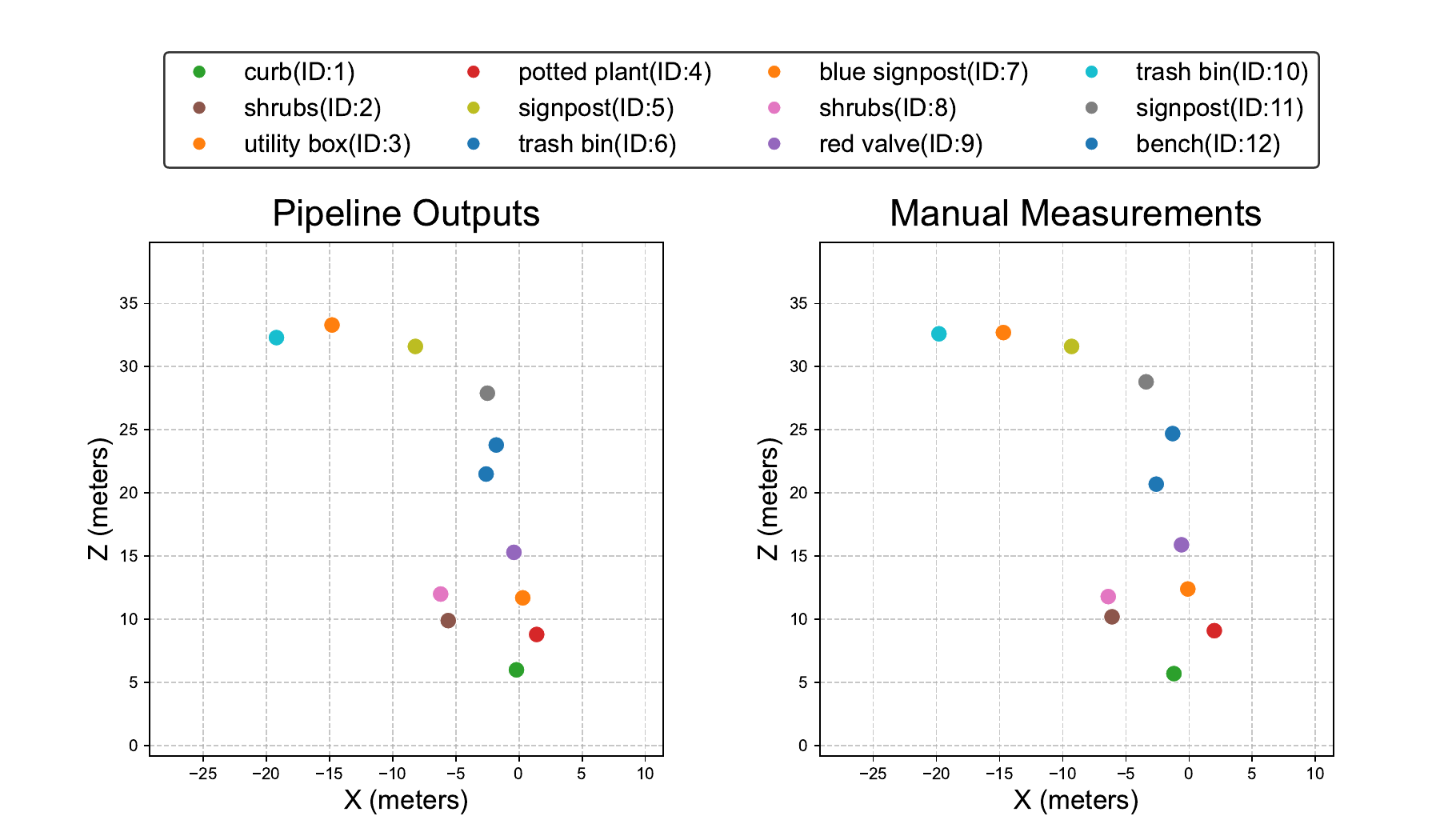}
        \caption{Validation of our pipeline in an outdoor campus scene.}
    \end{subfigure}
    \caption{\textbf{Qualitative Validation of the Pipeline.} We compare the map of static objects generated by our automated pipeline (left) against a ground-truth map of the same scene, which was measured manually on-site (right). The two maps are aligned for comparison.}
    \label{fig:error_map}
\end{figure}

To provide a qualitative validation of the pipeline's end-to-end accuracy, we conducted a real-world verification. We conducted this verification in two representative scenes: an indoor mall and an outdoor campus. For each scene, we first generated a 2D map of all static objects using our full pipeline. Subsequently, we returned to the physical locations to create a corresponding reference map by manually measuring the same objects' relative positions. As visualized in \cref{fig:error_map}, the comparison reveals a low mean positional error between the pipeline output and the manual ground truth: 0.68 meters for the indoor scene and 0.79 meters for the outdoor scene, respectively. This result confirms the high metric fidelity of our automated generation pipeline.

\section{Evaluation Details}
\label{sec:evaluation_detail}

\subsection{General Evaluation Setup}
All evaluations are conducted using the VLMEvalKit framework~\cite{duan2024vlmevalkit} to ensure a standardized protocol, running on NVIDIA RTX 4090 GPUs. Unless specified otherwise, we employ a greedy decoding strategy (i.e., temperature=0, top-k=1, top-p=1) for all models to ensure reproducibility. 

For closed-source models, we evaluate Gemini-2.5-Pro, Gemini-2.5-Flash~\cite{team2023gemini}, GPT-5, GPT-4o~\cite{hurst2024gpt4o}, Claude-3.7-Sonnet, Claude-4-Sonnet~\cite{Anthropic2024Claude}, Doubao-Seed-Vision~\cite{guo2025seed15vl}. For open-source models, we evaluate  Qwen2.5-VL~\cite{bai2025qwen25vl}, Qwen3-VL~\cite{bai2025qwen25vl}, InternVL2~\cite{chen2024internvl}, InternVL3.5~\cite{chen2024internvl}, LLaVA-OneVision~\cite{li2024llavaov}, LLaVA-Video~\cite{zhang2024llavanextvideo}, and Ovis2~\cite{lu2024ovis}, covering their variants with difference scales. 

The handling of the video frames modality varies by model. For all open-source models and the GPT series, we uniformly sample 32 frames per video as input. The Gemini-2.5-Pro model is the primary exception, as it supports native video ingestion, allowing us to provide the full MP4 file directly. See ~\cref{sec:effects_input_frames} for more detailed discussion for this.
Additionally, for models accessed via the OpenAI API (\eg GPT series), we enable the low-quality image mode, which forces each frame to be processed at a fixed cost of 85 tokens.

The input for all models follows a standard structure: ``[Pre-prompt][Question][Post-prompt][Video Frames]". The specific prompt templates for NA(Numerical Answers) and MCA(Multiple Choice Answers) are detailed in ~\cref{fig:prompts_main}. 

\begin{figure}
\begin{tcolorbox}[colback=black!5!white,colframe=black!75!black,title=Prompts]
\textbf{Pre-prompt}:\\
``These are frames of a video. In the video, objects are identified by numeric tags shown nearby. With that in mind, please answer the following question based on the video."\\
\textbf{Question}:\\
NA: question\_text.\\
MCA: question\_text + options.\\
\textbf{Post-prompt}:\\
NA: ``Your answer must be only the final numeric value, without units or any other text."\\
MCA: ``Your answer must be only the single letter (e.g., A, B, C, or D) of the correct option."
\end{tcolorbox}
\caption{\textbf{Prompts employed when constructing inputs for evaluations.}}
\label{fig:prompts_main}
\end{figure}

\subsection{Metrics and Baselines}
Although the prompt explicitly instructs the models to output only the final answer, we implement a regex-based fallback mechanism to handle non-compliant outputs. This parser extracts the last occurring numerical value or a valid multiple-choice option from the full generated text.

We employ two distinct metrics based on the answer format: standard \textit{Accuracy} ($\mathcal{ACC}$) for MCA questions, and \textit{Mean Relative Accuracy} ($\mathcal{MRA}$) for NA questions.
$\mathcal{MRA}$ measures the proportion of predefined error thresholds that the Mean Relative Error (MRE) can pass:
\begin{equation} \label{eq:mra}
    \mathcal{MRA} = \frac{1}{10} \sum_{e \in \mathcal{C}} \left( \frac{|\hat{y} - y|}{y} < e \right),
\end{equation}
where $\hat{y}$ and $y$ denote the prediction and ground truth, respectively, and $\mathcal{C}=\{0.05, 0.10, \dots, 0.50\}$ is the set of 10 error thresholds. This metric results in a step-wise score.

A special consideration is required for dynamic tasks where a near-zero ground truth makes the MRE denominator numerically unstable. We address this by defining a small threshold, $\mathcal{ST}$. For any ground truth $y < \mathcal{ST}$, a prediction $\hat{y} < \mathcal{ST}$ is awarded a full score of 1.0. If the prediction is incorrect ($\hat{y} \ge \mathcal{ST}$), the MRE is calculated with the denominator floored to $\mathcal{ST}$ to avoid division by zero. This threshold is set to a small value in practice, ensuring it only affects genuinely stationary or near-stationary cases.

For the four-choice MCA tasks, the chance-level baseline is 0.25. In contrast, we define the baseline for all NA tasks as \textit{zero}. This is because, for an uninformed guesser agnostic to the data distribution, the answer space is effectively unbounded—potentially ranging from centimeters to hundreds of meters—making any specific numerical guess fundamentally arbitrary. We argue that alternative baselines, such as the ``Frequency Chance Level" proposed by~\cite{yang2025thinking}, are invalid as they represent a data leak from the ground truth distribution.

\subsection{Human Evaluation Setup}
The human evaluation subset is a balanced collection of 270 questions, created by uniformly sampling 30 questions from each of our nine tasks. To ensure a consistent and high-quality evaluation, human annotators were given the instructions in ~\cref{fig:human_anno_instruct}.

\begin{figure*}
\begin{tcolorbox}[colback=black!5!white,colframe=black!75!black,title=Human Annotator Instructions]
\begin{itemize}
    \item Annotators are permitted unrestricted control over video playback. This includes the ability to play, pause, scrub the progress bar, and re-watch the video multiple times for each question to ensure their answer is as accurate as possible.
    \item The provided object captions (class names) may occasionally be imprecise. In cases of a conflict or discrepancy between the textual caption and the numerical ID tag shown in the video, the numerical ID should be considered the definitive ground truth. The object designated by the visual ID tag is the correct target for the question.
    \item For Numerical Answer (NA) tasks, provide the numerical value only, without any units (e.g., 12). For Multiple-Choice Answer (MCA) tasks, provide only the corresponding capital letter of the correct option (e.g., A).
    \item Before beginning the formal evaluation, annotators are provided with a calibration set. This set consists of 2 sample videos, their 20 corresponding question-answer pairs, and the associated ground truth (GT) answers. Annotators are instructed to review this material to familiarize themselves with the camera properties, the various question types, and to gain a reasonable sense of the metric scales used in the benchmark. This step ensures all annotators are properly calibrated before proceeding to the main evaluation tasks.
\end{itemize}
\end{tcolorbox}
\caption{\textbf{Human annotator instructions for evaluation.}}
\label{fig:human_anno_instruct}
\end{figure*}

Our human evaluation protocol follows the methodology of~\cite{yang2025thinking}, allowing annotators unrestricted control over video playback (e.g., play, pause, re-watch) to gather comprehensive information. Additionally, we employ a calibration phase (or ``warm-up" phase) specifically for open-world scenes, as we recognize the difficulty for human annotators to estimate real-world metric values based solely on visual inputs without training. Five human evaluators complete the entire subset independently and their scores are averaged to get the final human performance.

\subsection{Details of Synthetic Scenes.}
We generated our synthetic indoor data using the Blender Engine. First, we manually modeled 20 distinct indoor scenes spanning common layouts like washrooms, bedrooms, and living rooms. These scenes featured objects with conventional, real-world scales. We then leveraged the engine's internal metadata (e.g., object dimensions and locations) to automatically generate 120 template-based questions for our \textbf{Normal Set}, covering object size and distance tasks.
Next, we created a parallel \textbf{Abnormal Set} to serve as our testbed. For each of the 20 scenes, the object scales were deliberately manipulated to be counter-intuitive (e.g., a tiny bathtub, an oversized plant). Crucially, the overall scene layout and camera positions were kept identical to the Normal Set, isolating the variable of object scale. The same automated QA generation pipeline was then applied to this altered metadata to produce a corresponding set of 120 questions.

In the generated video, the camera is positioned at the room's center and performs two full 360-degree pans to ensure complete object coverage: the first pan is executed with a downward tilt, and the second with an upward tilt. Human evaluators for this test followed the same instructions as in the main evaluation.

\begin{figure}[t]
    \centering
    \begin{subfigure}{0.95\linewidth}
        \centering
        \includegraphics[width=\linewidth]{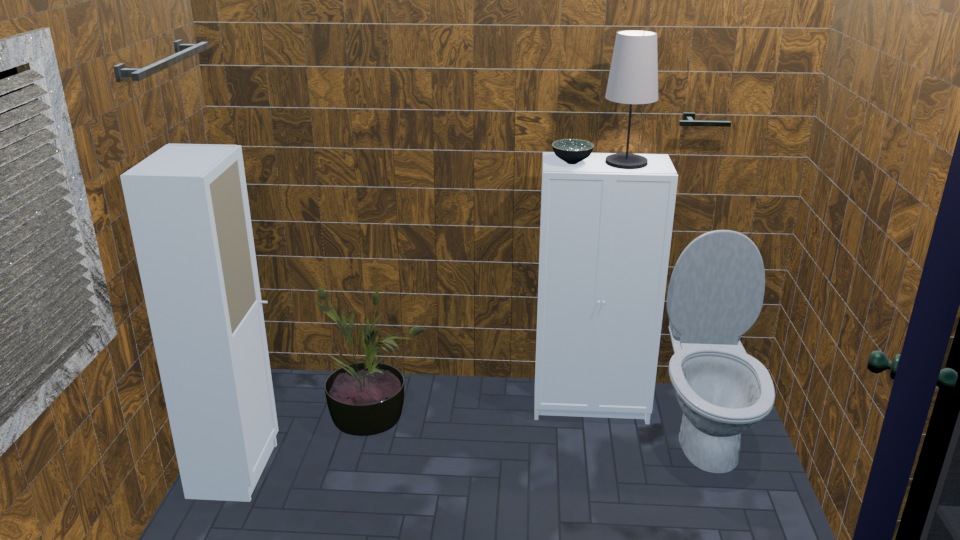}
        \caption{Normal scene.}
    \end{subfigure}
    \vspace{4pt}
    \begin{subfigure}{0.95\linewidth}
        \centering
        \includegraphics[width=\linewidth]{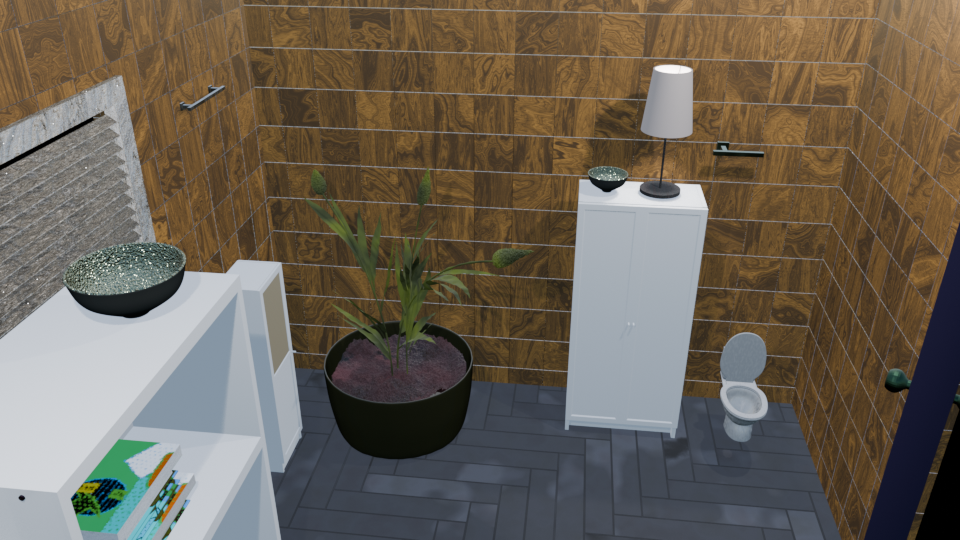}
        \caption{Abnormal scene. From the same perspective.}
    \end{subfigure}
    \caption{Samples from our synthetic test set.}
    \label{fig:synthetic_sample}
\end{figure}

See ~\cref{fig:synthetic_sample} for a sample of our synthetic test set and ~\cref{tab:synthetic_comparison} for detailed results.

\begin{table}[htbp]
    \centering
    \sisetup{detect-weight=true, detect-family=true} 

    \begin{threeparttable}
    \scalebox{0.74}{
    \begin{tabular}{
        l l  
        S[table-format=2.1]  
        S[table-format=2.1]  
        S[table-format=2.1, table-auto-round, round-mode=places, round-precision=1] 
    }
        \toprule
        \textbf{Model} & \textbf{Task} & {\textbf{Normal}} & {\textbf{Abnormal}} & {\textbf{Drop ($\Delta$)}} \\
        \midrule
        \multirow{3}{*}{Gemini-2.5-pro} 
         & Distance & 37.3 & 33.2 & 4.1 \\
         & Size     & 54.7 & 29.7 & 25.0 \\
         & \bfseries Overall & \bfseries 46.0 & \bfseries 31.4 & \bfseries 14.6 \\
        \midrule
        \multirow{3}{*}{Qwen2.5-VL-32B-Instruct}    
         & Distance & 43.0 & 33.0 & 10.0 \\
         & Size     & 54.5 & 28.3 & 26.2 \\
         & \bfseries Overall & \bfseries 48.8 & \bfseries 30.7 & \bfseries 18.1 \\
        \midrule
        \multirow{3}{*}{Human Performance}    
         & Distance & 51.7 & 51.4 & 0.3 \\
         & Size     & 62.5 & 60.5 & 2.0 \\
         & \bfseries Overall & \bfseries 57.1 & \bfseries 56.0 & \bfseries 1.1 \\
        \bottomrule
    \end{tabular}
    }
    \caption{Performance Degradation from Normal to Abnormal Conditions.}
    \label{tab:synthetic_comparison}
    \end{threeparttable}
\end{table}

\subsection{Details of Geometric Information Test}
To design this experiment, we first sampled \textit{absolute distance} questions from \Ourbench, filtering for cases where the two queried objects are not co-visible in any single frame. This selection criterion necessitates that the model reason using camera ego-motion, rather than simply calculating the distance between two objects in a static image. For this specific subset of questions, we then extracted the raw metadata ($p_1, p_2, t_1, t_2, R, T$) from our pipeline's intermediate outputs and formatted the tasks as shown in ~\cref{tab:abla_templates}.

\begin{table*}[h]
\centering
\begin{tabular}{p{4cm} p{10cm}}
\hline
\textbf{Component} & \textbf{Content} \\
\midrule
Question & What's the distance between the center of the \textcolor{red}{\{A\_class\}}(id: \textcolor{red}{\{A\_id\}}) and the \textcolor{red}{\{B\_class\}}(id: \textcolor{red}{\{B\_id\}}) in meters? \\
\hline
Formula post-prompt & In camera coordinates, x points right, y points down, and z points forward. To solve this, apply the following formula: $Distance = \| (R \cdot p_2 + T) - p_1 \|$. In this formula, $p_1$ is the 3D position in the camera coordinate of first queried object observed at the earlier time $t_1$, and $p_2$ is the 3D position of second queried object observed in the camera coordinate at the later time $t_2$. The matrix R and vector T represent the rotation and translation the camera pose has changed at time $t_2$ relative to time $t_1$. If any piece of information required to use the formula is not present in the text, you must infer it from the video and then use it in the formula.\\
\hline
Obj1 Info & At \textcolor{red}{\{t1\}}s($t_1$) of the video, \textcolor{red}{\{B\_class\}}(id: \textcolor{red}{\{B\_id\}}) is located at $p_1$ = [\textcolor{red}{\{p1\_x\}},\textcolor{red}{\{p1\_y\}},\textcolor{red}{\{p1\_z\}}] meters relative to the camera.\\
\hline
Obj2 Info & At \textcolor{red}{\{t2\}}s($t_2$) of the video, \textcolor{red}{\{A\_class\}}(id: \textcolor{red}{\{A\_id\}}) is located at $p_2$ = [\textcolor{red}{\{p2\_x\}},\textcolor{red}{\{p2\_y\}},\textcolor{red}{\{p2\_z\}}] meters relative to the camera.\\
\hline
Ego-motion Info & Between \textcolor{red}{\{t1\}}s($t_1$) and \textcolor{red}{\{t2\}}s($t_2$), the camera's relative translation is $T$ = [\textcolor{red}{\{T\_x\}}, \textcolor{red}{\{T\_y\}}, \textcolor{red}{\{T\_z\}}] and the rotation matrix is $R =$ [[\textcolor{red}{\{R11\}}, \textcolor{red}{\{R12\}}, \textcolor{red}{\{R13\}}], [\textcolor{red}{\{R21\}}, \textcolor{red}{\{R22\}}, \textcolor{red}{\{R23\}}], [\textcolor{red}{\{R31\}}, \textcolor{red}{\{R32\}}, \textcolor{red}{\{R33\}}]].\\
\bottomrule
\end{tabular}
\begin{tabular}{p{4cm} p{10cm}}
\toprule
\textbf{Setting} & \textbf{Question Template} \\
\midrule
Vanilla & 
\{Question\} \{Formula post-prompt\} \\
\hline
+ One Localization($p_1$) & 
\{Obj1 Info\} \{Question\} \{Formula post-prompt\} \\
\hline
+ Both Localization($p_1$,$p_2$) &
\{Obj1 Info\} \{Obj2 Info\} \{Question\} \{Formula post-prompt\} \\
\hline
+ Ego-Motion($R,T$) &
\{Ego-motion Info\} \{Question\} \{Formula post-prompt\} \\
\hline
+ All($p_1, p_2, R, T$) &
\{Obj1 Info\} \{Obj2 Info\} \{Ego-motion Info\} \{Question\} \{Formula post-prompt\} \\
\hline
+ All(w/o Formula) & \{Obj1 Info\} \{Obj2 Info\} \{Ego-motion Info\} \{Question\}\\
\bottomrule
\end{tabular}
\caption{\textbf{Templates for the Geometric Information Test.}}
\label{tab:abla_templates}
\end{table*}

\section{More Results}
\subsection{Full Evaluation Results}

\begin{figure*}[b]
    \captionsetup{type=table}
    \vspace{-0.4cm}
    \centering
    \fontsize{10.2pt}{10.0pt}\selectfont
    \setlength\tabcolsep{4pt}  
    \renewcommand{\arraystretch}{1.0}  
    \scalebox{0.8}{
    \begin{tabular}{r|cc|ccccccccc} 
    & & & 
    \rot{Rel. Dis.} &
    \rot{Rel. Dir.} &
    \rot{Qual. S-Motion} & 
    \rot{Obj. Loc.} &
    \rot{Abs. Dis.} &
    \rot{Depth Count} &
    \rot{Abs. Displ.} &
    \rot{Abs. Speed} &
    \rot{Quan. S-Motion}
    \\
    Methods & Rank & Avg. & \multicolumn{3}{c}{\cellcolor{green!20}Relational$(\mathcal{MCA})$} & \multicolumn{3}{c}{\cellcolor{purple!20}Static Metric$(\mathcal{NA})$} & \multicolumn{3}{c}{\cellcolor{pink!20}Dynamic Metric$(\mathcal{NA})$}\\
    \hline

    \rowcolor{green!10}
    \multicolumn{1}{l|}{\textcolor{black}{\textit{Against Human on tiny}}}  & & & & & & & & & & &  \\
    Human-level & - & 60.3 & 85.7 & 83.3 & 73.7 & 43.9 & 39.2 & 67.5 & 42.9 & 65.8 & 66.8 \\
    Gemini-2.5-Pro & - & 36.8 & 53.1 & 23.1 & 46.7 & 39.7 & 33.8 & 40.3 & 22.2 & 27.8 & 40.0 \\
    GPT-5 & - & 27.9 & 37.5 & 30.8 & 40.0 & 35.3 & 25.3 & 12.8 & 9.2 & 31.4 & 33.0 \\
    Qwen2.5VL-32B-Instruct & - & 32.1 & 68.8 & 23.1 & 33.3 & 14.4 & 29.7 & 31.3 & 17.5 & 32.8 & 35.7 \\
    \hline
    
    \rowcolor{green!10}
    \multicolumn{1}{l|}{\textcolor{black}{\textit{Closed-source Models}}}  & & & & & & & & & & &  \\
    Gemini-2.5-Pro & - & 37.2 & 50.0 & 28.1 & 52.5 & 37.4 & 28.1 & 37.9 & 26.8 & 31.1 & 40.8 \\
    Gemini-2.5-Flash & - & 19.5 & 17.9 & 2.8 & 50.6 & 22.7 & 16.1 & 26.8 & 8.1 & 6.8 & 20.0 \\
    GPT-5 & - & 29.7 & 34.4 & 33.1 & 49.5 & 32.5 & 23.7 & 20.9 & 10.5 & 33.8 & 30.6\\
    GPT-4o & - & 25.9 & 30.8 & 29.1 & 42.2 & 22.9 & 27.0 & 21.6 & 17.5 & 15.5 & 28.8 \\
    Claude-3.7-Sonnet & - & 26.5 & 38.9 & 32.8 & 47.6 & 31.3 & 22.4 & 31.5 & 5.2 & 30.1 & 5.0 \\
    Doubao-Seed-1.6V & - & 27.3 & 35.9 & 24.1 & 44.0 & 16.6 & 18.9 & 38.7 & 25.7 & 31.8 & 9.2 \\
    Claude-4-Sonnet & - & 13.2 & 7.2 & 0.7 & 2.8 & 20.0 & 21.6 & 26.4 & 9.0 & 17.9 & 9.0 \\
    Grok5(8f) & - & 13.6 & 23.5 & 19.6 & 2.1 & 10.1 & 10.5 & 24.2 & 12.8 & 13.9 & 7.9 \\
    \hline

    \rowcolor{green!10}
    \multicolumn{1}{l|}{\textcolor{black}{\textit{Open-source Models}}}  & & & & & & & & & & &  \\
    InternVL2-1B & - & 19.6 & 33.2 & 32.1 & 40.8 & 7.6 & 15.4 & 21.8 & 7.1 & 14.3 & 8.8 \\
    InternVL2-2B & - & 17.1 & 30.4 & 18.6 & 40.6 & 7.0 & 7.8 & 32.9 & 4.7 & 13.2 & 0.1 \\
    InternVL2-4B & - & 21.7 & 27.2 & 28.1 & 40.6 & 14.6 & 21.0 & 36.9 & 12.2 & 17.2 & 0.0 \\
    InternVL2-8B & - & 24.5 & 35.1 & 31.7 & 40.8 & 21.8 & 17.8 & 39.7 & 15.0 & 17.8 & 3.8 \\
    InternVL2-26B & - & 26.0 & 34.5 & 34.0 & 40.7 & 28.9 & 20.6 & 38.3 & 12.3 & 23.6 & 4.6 \\
    InternVL2-40B & - & 22.9 & 36.7 & 21.0 & 41.9 & 21.1 & 19.2 & 33.0 & 10.0 & 22.0 & 1.7 \\
    InternVL2-76B & - & 25.5 & 32.3 & 30.6 & 41.3 & 15.8 & 13.5 & 39.9 & 15.3 & 25.0 & 18.2 \\
    InternVL3.5-1B & - & 19.0 & 33.4 & 33.0 & 36.5 & 2.4 & 1.0 & 28.1 & 7.0 & 14.8 & 20.8 \\
    InternVL3.5-2B & - & 21.7 & 34.8 & 32.1 & 40.1 & 3.8 & 2.7 & 40.8 & 11.5 & 16.6 & 17.0 \\
    InternVL3.5-4B & - & 23.7 & 37.6 & 32.8 & 44.8 & 4.6 & 6.7 & 40.7 & 15.6 & 23.4 & 10.7 \\
    InternVL3.5-8B & - & 28.5 & 37.6 & 33.6 & 47.3 & 12.2 & 13.2 & 42.3 & 20.3 & 30.2 & 21.5 \\
    InternVL3.5-14B & - & 28.5 & 40.3 & 33.9 & 47.1 & 15.5 & 15.6 & 42.8 & 21.8 & 32.1 & 9.0 \\
    InternVL3.5-38B & - & 26.9 & 40.2 & 34.0 & 45.3 & 11.6 & 7.7 & 42.7 & 20.3 & 31.4 & 11.1 \\
    Qwen2.5VL-3B-Instruct & - & 24.2 & 30.3 & 32.7 & 43.0 & 17.2 & 26.1 & 18.4 & 13.3 & 19.1 & 21.1 \\
    Qwen2.5VL-7B-Instruct & - & 27.1 & 33.1 & 17.3 & 41.5 & 22.0 & 25.1 & 27.6 & 18.3 & 26.5 & 30.0 \\
    Qwen2.5VL-32B-Instruct & - & 30.0 & 41.7 & 32.7 & 44.4 & 23.9 & 24.0 & 27.7 & 16.7 & 32.8 & 27.3 \\
    Qwen2.5VL-72B-Instruct & - & 26.5 & 38.5 & 20.1 & 46.5 & 27.7 & 26.0 & 29.4 & 9.4 & 29.7 & 9.8 \\
    Qwen3VL-2B-Instruct & - & 18.4 & 33.7 & 32.4 & 37.5 & 2.2 & 4.2 & 22.4 & 6.6 & 19.0 & 12.5 \\
    Qwen3VL-4B-Instruct & - & 21.1 & 34.8 & 23.7 & 46.4 & 3.9 & 6.9 & 24.1 & 13.6 & 20.4 & 17.5 \\
    Qwen3VL-8B-Instruct & - & 31.2 & 38.3 & 31.2 & 49.3 & 21.0 & 15.1 & 33.3 & 21.3 & 34.3 & 37.8 \\
    Qwen3VL-32B-Instruct & - & 32.2 & 41.9 & 28.8 & 47.1 & 25.3 & 11.5 & 30.2 & 18.6 & 36.8 & 49.2 \\
    Qwen3VL-30B-A3B-Instruct & - & 29.3 & 37.5 & 29.9 & 44.0 & 12.4 & 8.7 & 33.1 & 20.4 & 31.7 & 46.9 \\
    LLaVA-OneVision-0.5B & - & 19.6 & 27.9 & 23.8 & 40.8 & 13.5 & 13.1 & 19.9 & 10.3 & 13.6 & 15.7 \\
    LLaVA-OneVision-7B & - & 25.7 & 35.1 & 32.7 & 40.8 & 16.1 & 25.5 & 25.6 & 16.8 & 28.3 & 13.4 \\
    LLaVA-OneVision-72B & - & 26.9 & 38.6 & 32.2 & 41.7 & 19.6 & 18.3 & 35.3 & 19.2 & 23.0 & 16.6 \\
    LLaVA-Video-Qwen2-7B & - & 22.9 & 37.1 & 31.2 & 40.9 & 17.6 & 22.1 & 18.2 & 17.5 & 19.0 & 5.7 \\
    LLaVA-Video-Qwen2-72B & - & 28.3 & 39.8 & 31.1 & 42.2 & 23.5 & 18.0 & 34.2 & 20.7 & 29.9 & 17.0 \\
    Ovis2-4B & - & 25.9 & 34.6 & 30.3 & 40.1 & 16.1 & 18.6 & 36.8 & 17.7 & 22.6 & 18.7 \\
    Ovis2-16B & - & 28.1 & 37.0 & 35.8 & 42.0 & 20.0 & 6.2 & 41.7 & 21.7 & 23.3 & 28.2 \\
    Ovis2-34B & - & 26.8 & 37.3 & 35.5 & 40.8 & 19.1 & 13.1 & 37.5 & 18.7 & 27.4 & 15.5 \\
    \end{tabular}
    }
    \vspace{-0.2cm}
    \caption{\textbf{Full Evaluation results for all MLLMs we tested.} }
    \label{tab:main_table_full_results}
\end{figure*}

In addition to the key results shown in the main paper, we evaluated other models such as the full InternVL2 series~\cite{chen2024internvl} and Qwen2.5VL series~\cite{bai2025qwen25vl}. We provide the comprehensive results for all generalist models tested in our study in ~\cref{tab:main_table_full_results}.

During this extensive evaluation, we found that two models, Claude-4-Sonnet and Grok5 (with 8-frame input), exhibited collapsed performance. This was due to a consistent failure to adhere to the task's prompt instructions, rather than a specific failure in spatial reasoning. 

\subsection{Effects of Input Frames}
\label{sec:effects_input_frames}
For all evaluated models, with the exception of the Gemini family, we uniformly sample 32 frames per video as image inputs. These frames are presented chronologically, and the prompt is augmented with temporal information (e.g., video duration and number of sampled frames) to enable time-related reasoning. The Gemini family is the primary exception, as it supports native video ingestion; we provide the full MP4 file directly, as we consider this an integral part of its capability. A second exception is Grok-5, which was limited to 8 frames due to the constraints of the API we used.

We then conducted an ablation study to quantify the impact of frame count on performance, using Qwen3VL-32B (the top-performing open-source model) as our testbed. The results, shown in \cref{tab:frames_ablation}, indicate that while increasing the number of sampled frames yields a slight performance benefit, the overall gain is marginal.

\begin{figure*}[ht!]
    \captionsetup{type=table}
    \centering
    \fontsize{10.2pt}{10.0pt}\selectfont
    \setlength\tabcolsep{4pt}  
    \renewcommand{\arraystretch}{1.0}  
    \scalebox{0.87}{
    \begin{tabular}{c|c|ccccccccc} 
    & & 
    \rot{Rel. Dis.} &
    \rot{Rel. Dir.} &
    \rot{Qual. S-Motion} & 
    \rot{Obj. Loc.} &
    \rot{Abs. Dis.} &
    \rot{Depth Count} &
    \rot{Abs. Displ.} &
    \rot{Abs. Speed} &
    \rot{Quan. S-Motion}
    \\
    Num. of Frames & Avg. & \multicolumn{3}{c}{\cellcolor{green!20}Relational} & \multicolumn{3}{c}{\cellcolor{purple!20}Static Metric} & \multicolumn{3}{c}{\cellcolor{pink!20}Dynamic Metric}\\
    \hline
    8  & 30.2 & 40.6 & 29.7 & 47.4 & 21.9 & 12.6 & 29.2 & 18.5 & 32.0 & 40.5 \\
    16 & 31.8 & 42.2 & 29.7 & 47.3 & 23.3 & 11.5 & 31.0 & 20.5 & 35.6 & 45.1 \\
    32 & 32.2 & 41.9 & 28.8 & 47.1 & 25.3 & 11.5 & 30.2 & 18.6 & 36.8 & 49.2 \\
    64 & 32.5 & 42.8 & 30.2 & 47.0 & 26.6 & 10.4 & 31.6 & 17.5 & 36.5 & 50.2 \\
    \end{tabular}
    }
\caption{\textbf{Ablation study on the number of sampled input frames for Qwen3-VL-32B.}}
\label{tab:frames_ablation}
\end{figure*}

\subsection{Results of Spatial Models}
As a comparison, we also evaluated three specialized spatial models: SpatialRGPT-VILA1.5-8B~\cite{cheng2024spatialrgpt}, Space-Thinker-Qwen, and SpaceOm~\cite{chen2024spatialvlm} (the latter two finetuned on Qwen2.5-VL-3B). The results are presented in \cref{tab:spatial_models}. We warn, however, that these scores are not a direct, like-for-like comparison due to fundamental misalignments in task design. SpatialRGPT is a region-prompted model, and to adapt it to our whole-video tasks, we provided a full-image mask as input to give access for the whole depth map to the model. Similarly, SpaceThinker and SpaceOm were not specifically trained for video-based reasoning, which may explain their performance relative to their base models. Therefore, these results are reported primarily as an initial reference.

\begin{figure*}[ht!]
    \captionsetup{type=table}
    \centering
    \fontsize{10.2pt}{10.0pt}\selectfont
    \setlength\tabcolsep{4pt}  
    \renewcommand{\arraystretch}{1.0}  
    \scalebox{0.85}{
    \begin{tabular}{c|c|ccccccccc} 
    & & 
    \rot{Rel. Dis.} &
    \rot{Rel. Dir.} &
    \rot{Qual. S-Motion} & 
    \rot{Obj. Loc.} &
    \rot{Abs. Dis.} &
    \rot{Depth Count} &
    \rot{Abs. Displ.} &
    \rot{Abs. Speed} &
    \rot{Quan. S-Motion}
    \\
    Model & Avg. & \multicolumn{3}{c}{\cellcolor{green!20}Relational} & \multicolumn{3}{c}{\cellcolor{purple!20}Static Metric} & \multicolumn{3}{c}{\cellcolor{pink!20}Dynamic Metric}\\
    \hline
VILA-1.5-8B  & 20.4 & 30.5 & 30.0 & 40.8 & 17.7 & 25.2 & 17.1 & 15.9 & 8.8 & 1.3 \\
SpatialRGPT-VILA-1.5-8B  & 24.0(+3.6) & 31.9 & 33.0 & 40.8 & 23.1 & 16.1 & 26.0 & 19.5 & 13.9 & 15.5 \\
Qwen2.5VL-3B-Instruct & 24.2 & 30.3 & 32.7 & 43.0 & 17.2 & 26.1 & 18.4 & 13.3 & 19.1 & 21.1 \\
SpaceThinker-Qwen-3B  & 21.7(-2.5) & 29.3 & 31.9 & 40.8 & 0.2 & 0.3 & 38.4 & 6.2 & 28.0 & 23.6 \\
SpaceOm-3B & 23.2(-1.0) & 29.8 & 26.3 & 44.0 & 13.0 & 23.4 & 26.5 & 16.1 & 22.6 & 8.5 \\
    \end{tabular}
}
\caption{\textbf{Results for Specialized Spatial Models and Their Corresponding Base Model.}}
\label{tab:spatial_models}
\end{figure*}

\subsection{Detailed Results of the Comparison for Model Generations}
See \cref{tab:generation_comparison_openbench} and \cref{tab:generation_comparison_vsi} for the detailed results comparing the QwenVL and InternVL families across different sizes and generations on both \Ourbench and VSI-Bench. While a direct, like-for-like comparison at each model size is constrained by model availability, two primary findings emerge from this data: (i) on \Ourbench, the relationship between performance and model size is non-monotonic, often saturating or degrading; and (ii) the performance gain from newer generations on \Ourbench is marginal, which stands in stark contrast to the significant and stable gains reported on VSI-Bench (\eg \textgreater +23.0 for both families and all sizes).

Our main paper's analysis focuses on the comparison between \Ourbench and VSI-Bench~\cite{yang2025thinking}, as their shared video modality and tasks allow for a direct comparison. To validate that our findings regarding illusory progress on indoor benchmarks are not an artifact of VSI-Bench alone, we conducted an additional test on the multi-view indoor spatial benchmark, All-Angle-Bench~\cite{yeh2025seeingperspective}. The results in ~\cref{tab:generation_comparison_all_angles} confirm that the trend of performance gains from larger model scales and newer model generations persists on this benchmark as well. However, the magnitude of the gain from newer model generations is less significant than that observed on VSI-Bench. We hypothesize that this reduced gain may be due to the models' lack of specific training for the multi-view data format required by All-Angle-Bench.

\begin{table}[ht]
    \centering
    \scalebox{0.85}{
    \begin{threeparttable} 
        \small 
        \begin{tabular}{l S S S S}
            \toprule
            & {\textbf{InternVL2}} & {\textbf{InternVL3.5}} & {\textbf{QwenVL2.5}} & {\textbf{QwenVL3}} \\
            \midrule
            2B                & 21.7 & 21.7 & {---} & 17.1 \\
            3B/4B             & 21.7 & 23.7 & 24.2  & 21.1 \\
            7B/8B             & 24.5 & 28.5 & 27.1  & 31.2 \\
            14B/26B           & 26.0 & 28.5 & {---} & {---} \\
            32B\tnote{1}      & 22.9 & 26.9 & 30.0  & 32.2 \\ 
            72B/76B           & 25.5 & {---} & 26.5  & {---} \\
            \bottomrule
        \end{tabular}
        \begin{tablenotes}[flushleft]
            \footnotesize
            \item[1] This line also includes 38B/40B for InternVL series.
        \end{tablenotes}
    \caption{\textbf{Comparisons of models on \Ourbench across size and generations.}}
    \label{tab:generation_comparison_openbench}
    \end{threeparttable}
    }
\end{table}

\begin{table}[ht]
    \centering
    \scalebox{0.85}{
    \begin{threeparttable} 
        \small 
        \begin{tabular}{l S S S S}
            \toprule
            & {\textbf{InternVL2}} & {\textbf{InternVL3.5}} & {\textbf{QwenVL2.5}} & {\textbf{QwenVL3}} \\
            \midrule
            2B                & 26.7 & 52.8 & {---} & 53.9 \\
            3B/4B             & 33.5 & 56.6 & 27.9  & 58.4 \\
            7B/8B             & 37.6 & 56.1 & 36.8  & 59.4 \\
            32B\tnote{1}      & 37.3 & 61.4 & 37.7 & 61.5 \\ 
            \bottomrule
        \end{tabular}
        \begin{tablenotes}[flushleft]
            \footnotesize
            \item[1] This line also includes 38B/40B for InternVL series.
        \end{tablenotes}
    \caption{\textbf{Comparisons of models on VSI-Bench~\cite{yang2025thinking} across size and generations.}}
    \label{tab:generation_comparison_vsi}
    \end{threeparttable}
    }
\end{table}

\begin{table}[ht]
    \centering
    \scalebox{0.85}{
    \begin{threeparttable} 
        \small 
        \begin{tabular}{l S S S S}
            \toprule
            & {\textbf{InternVL2}} & {\textbf{InternVL3.5}} & {\textbf{QwenVL2.5}} & {\textbf{QwenVL3}} \\
            \midrule
            2B                & 40.9 & 45.9 & {---} & 44.6 \\
            3B/4B             & 43.2 & 49.0 & 42.7  & 49.2 \\
            7B/8B             & 47.7 & 52.1 & 48.5  & 50.9 \\
            14B/26B           & 50.3 & 51.0 & {---} & {---} \\
            32B\tnote{1}      & 50.8 & 53.7 & 54.6  & 55.7 \\ 
            72B/76B           & 50.9 & {---} & {55.1}  & {---} \\
            \bottomrule
        \end{tabular}
        \begin{tablenotes}[flushleft]
            \footnotesize
            \item[1] This line also includes 38B/40B for InternVL series.
        \end{tablenotes}
    \caption{\textbf{Comparisons of models on ALL-Angles-Bench~\cite{yeh2025seeingperspective} across size and generations.}}
    \label{tab:generation_comparison_all_angles}
    \end{threeparttable}
    }
\end{table}

\section{Privacy}
To address privacy concerns, our data acquisition was conducted exclusively in publicly accessible locations. Subsequently, all collected data underwent a rigorous human curation process to identify and remove any potentially sensitive or private information.

\end{document}